\PassOptionsToPackage{table}{xcolor}
\documentclass[a4paper,fleqn]{cas-sc}
\usepackage[authoryear,round]{natbib}
\usepackage[utf8]{inputenc}
\usepackage{tabularx}
\usepackage{subcaption}
\usepackage{placeins}
\usepackage{microtype}

\usepackage{xurl}

\definecolor{hanpurple}{rgb}{0.32,0.09,0.98}
\definecolor{mygray}{gray}{.9}

\newcolumntype{Y}{>{\raggedright\arraybackslash}X}

\ExplSyntaxOn
\NewDocumentCommand{\LocalFigureWidth}{}{
  \dim_set:Nn \l_fig_width_dim {\linewidth}
}
\RenewDocumentCommand{\printorcid}{}{
  \seq_if_empty:NF \g_stm_orcid_seq {
    \group_begin:
    \renewcommand\thefootnote{}
    \footnotetext{\textsc{orcid}(s):~\seq_use:Nn \g_stm_orcid_seq {;~}}
    \group_end:
  }
}
\ExplSyntaxOff

\newif\ifshowhighlights
\showhighlightsfalse

\begin{document}
\shorttitle{EgoEV-HandPose}
\shortauthors{Wang et al.}
\title[mode=title]{EgoEV-HandPose: Egocentric 3D Hand Pose Estimation and Gesture Recognition with Stereo Event Cameras}

\author[1]{Luming Wang}
\author[2]{Hao Shi}\cormark[2]\ead{sh531467@antgroup.com}
\author[1]{Jiajun Zhai}
\author[3]{Kailun Yang}
\author[1,4]{Kaiwei Wang}\cormark[1]\ead{wangkaiwei@zju.edu.cn}

\affiliation[1]{organization={National Research Center for Optical Instrumentation, Zhejiang University},
  city={Hangzhou}, postcode={310027}, country={China}}
\affiliation[2]{organization={Ant Group Company Ltd.},
  city={Hangzhou}, postcode={310013}, country={China}}
\affiliation[3]{organization={School of Artificial Intelligence and Robotics, and National Engineering Research Center of Robot Visual Perception and Control Technology, Hunan University},
  city={Changsha}, postcode={410082}, country={China}}
\affiliation[4]{organization={Central Research Institute of Sunny Optical Technology},
  city={Hangzhou}, postcode={311215}, country={China}}

\cortext[1]{Corresponding author: Kaiwei Wang (wangkaiwei@zju.edu.cn).}
\cortext[2]{Corresponding author: Hao Shi (sh531467@antgroup.com).}
\nonumnote{This research was funded by the Natural Science Foundation of Zhejiang Province (Grant No. LZ24F050003), the National Natural Science Foundation of China (Grant No. 62473139), the Hunan Provincial Research and Development Project (Grant No. 2025QK3019), and the opening project of the State Key Laboratory of Autonomous Intelligent Unmanned Systems (Grant No. ZZKF2025-2-10).}

\begin{abstract}
Egocentric 3D hand pose estimation and gesture recognition are fundamental to understanding hand interactions from first-person observations. Robust perception remains challenging: frame-based cameras suffer from motion blur and limited dynamic range, while event-based approaches face ego-motion interference, monocular depth ambiguity, and limited real-world datasets. We propose EgoEV-HandPose, an end-to-end framework for joint 3D bimanual hand pose estimation and gesture recognition from stereo event streams. Its core module, KeypointBEV, combines sparse 3D joint hypotheses with geometry-aware stereo fusion and iterative reprojection-guided refinement to progressively resolve depth uncertainty and improve pose accuracy. We also introduce EgoEVHands, the first large-scale real-world stereo event dataset for egocentric 3D bimanual hand pose estimation and gesture recognition. Built upon EgoEvGesture recordings, it adds dense 2D/3D hand annotations and comprises 2,498 annotated stereo sequence pairs across 38 gesture classes. Extensive experiments demonstrate that EgoEV-HandPose achieves state-of-the-art performance on EgoEVHands, obtaining a mean per-joint position error (MPJPE) of $30.54\,\mathrm{mm}$ and a gesture recognition accuracy of $86.87\%$, significantly outperforming RGB-based stereo and previous event-based methods. The framework exhibits exceptional robustness under low illumination and severe bimanual occlusion. Experiments on DHP19 further demonstrate the cross-dataset and cross-architecture generalizability of the KeypointBEV refinement design. Dataset and code will be released at \url{https://github.com/ZJUWang01/EgoEV-HandPose}.
\end{abstract}


\maketitle

\begin{figure}[pos=!htbp,width=\linewidth,align=\centering,abovecap=4pt,belowcap=2pt]
\includegraphics[width=\linewidth,height=.22\textheight,keepaspectratio]{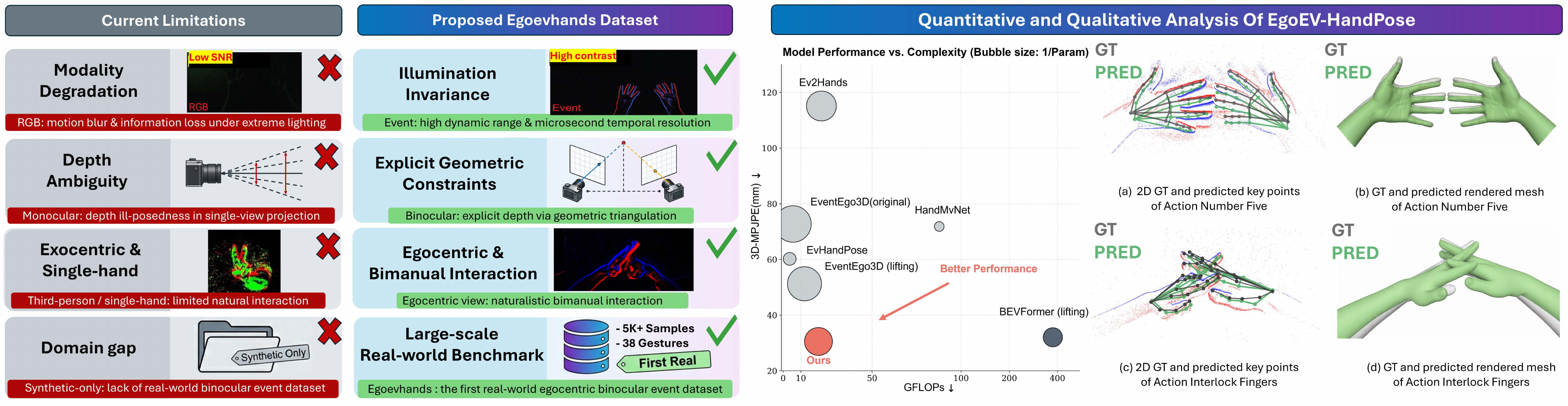}
  \caption{Overview of the proposed EgoEV-HandPose framework and the established EgoEVHands dataset.
  Left: challenges in frame-based and monocular hand perception. Middle: EgoEVHands builds on EgoEvGesture recordings to establish the first large-scale real-world stereo event dataset for egocentric 3D bimanual hand pose estimation and gesture recognition, with dense pose-level annotations. Right: state-of-the-art pose accuracy and a favorable accuracy--complexity trade-off.}
  \label{fig:fig1}
\end{figure}

\section{Introduction}
Egocentric 3D hand pose estimation and gesture recognition connect the geometric reconstruction of articulated hands with the interpretation of human actions. They support human--computer interaction~\citep{Zhang2018EgoGesture}, articulated hand tracking~\citep{han2020megatrack}, and the visual analysis of everyday activities~\citep{Grauman2022Ego4D,Sener2022Assembly101}. The egocentric viewpoint aligns observations with the user's actions, but recovering two interacting hands from a moving camera remains challenging because of self-occlusion, rapid motion, background-motion interference, changing illumination, and uncertain depth. Conventional frame-based solutions are further limited by motion blur and restricted dynamic range, which degrade the visual observations needed for hand localization~\citep{Grauman2022Ego4D,Garcia-Hernando2018FPHA,chi2023egovsr}. Event cameras offer asynchronous sensing with high temporal resolution and a wide dynamic range~\citep{Gallego2022Event}, making them well suited to capturing hand dynamics under challenging motion and illumination conditions.

Despite these sensing advantages, event-based egocentric hand perception still faces geometric and data challenges. Monocular event observations leave depth recovery ambiguous, particularly during occluded bimanual interactions. Parametric hand models such as MANO provide useful priors~\citep{Rudnev2021EventHands,Jiang2024EvHandPose,Millerdurai2024Ev2Hands}, but do not replace explicit geometric constraints. Meanwhile, head motion generates background events that can interfere with the hand observations~\citep{Hara2025EventEgoHands,millerdurai2024eventego3d}. Real-world stereo event recordings with pose-level annotations also remain limited, restricting the development and evaluation of geometry-aware bimanual perception.

We address these challenges with EgoEV-HandPose, an end-to-end framework for 3D bimanual hand pose estimation and gesture recognition (Fig.~\ref{fig:fig1}). A shared EgoBlaze backbone extracts features and 2D keypoints from the stereo event observations. The proposed KeypointBEV module uses an analytical stereo initialization and iteratively refines sparse 3D joint hypotheses through reprojection-guided feature sampling. By combining joint-specific visual evidence with global stereo context, it exploits cross-view geometry without constructing a dense 3D voxel grid. A temporal classifier interprets the refined hand trajectories for gesture recognition within the jointly optimized framework. To support geometry-aware learning and evaluation, we also introduce EgoEVHands, the first large-scale real-world stereo event dataset for egocentric 3D bimanual hand pose estimation and gesture recognition. Built upon the stereo recordings originally collected for EgoEvGesture~\citep{Wang2025EgoEvGesture}, it retains the recording protocol and gesture labels while adding comprehensive pose-level supervision. We construct dense 2D/3D hand keypoints, cross-view projections, segmentation masks, and joint-level validity information, transforming the original gesture-recognition resource into a unified dataset for stereo geometric reconstruction and hand interaction understanding.

Extensive experiments demonstrate that EgoEV-HandPose achieves state-of-the-art performance on EgoEVHands, reaching $30.54$\,mm MPJPE and $86.87\%$ gesture recognition accuracy in a cross-subject setting where the test participants are excluded from training. The pose error is $57.5\%$ lower than that of HandMvNet~\citep{Ali2025HandMvNet} and $73.5\%$ lower than that of Ev2Hands~\citep{Millerdurai2024Ev2Hands}. Fixed-backbone comparisons and fusion ablations demonstrate the effectiveness of KeypointBEV's geometry-guided stereo fusion and iterative refinement in improving 3D pose accuracy, while the scenario-wise results demonstrate exceptional robustness to low illumination and bimanual occlusion. Beyond hand perception, integrating KeypointBEV with DEV-Pose~\citep{yin2023rethinking} on DHP19~\citep{calabrese2019dhp19} reduces MPJPE from $55.53$ to $47.88$\,mm, demonstrating the plug-and-play versatility and cross-dataset generalizability of the geometric refinement design.

The contributions are threefold: (i) EgoEVHands, the first large-scale real-world stereo event dataset for egocentric 3D bimanual hand pose estimation and gesture recognition, extending EgoEvGesture recordings with dense pose annotations and comprising 2,498 annotated stereo sequence pairs across 38 gesture categories under both normal-light and low-light conditions; (ii) EgoEV-HandPose, an end-to-end framework whose KeypointBEV module anchors sparse 3D joint hypotheses to stereo observations through reprojection-guided fusion and iterative refinement, enabling accurate geometric reconstruction and discriminative gesture understanding; and (iii) extensive evaluations demonstrating state-of-the-art performance, strong cross-subject generalization, and exceptional robustness under challenging illumination and occlusion, together with cross-dataset and cross-architecture experiments on DHP19 that establish the plug-and-play effectiveness of KeypointBEV.

\section{Related Work}
\label{sec:related_work}

\subsection{Frame-based Egocentric Hand Perception}
\label{subsec:hand_analysis}

Egocentric hand perception requires estimating articulated geometry and interpreting interactions under severe self-occlusion and motion blur~\citep{Grauman2022Ego4D, Garcia-Hernando2018FPHA}. Recent progress in frame-based methods has been fueled by large-scale egocentric datasets~\citep{Grauman2022Ego4D, Kwon2021H2O, Sener2022Assembly101, Garcia-Hernando2018FPHA, Fan2023ARCTIC, Ohkawa2023AssemblyHands, Lin2024Ego2HandsPose, Zhang2018EgoGesture}.
Early monocular 3D hand pose estimation employed direct keypoint regression~\citep{Zimmermann2017Learning, Doosti2020HOPE} for its lightweight skeletal output, yet it is fundamentally hindered by ill-posed depth ambiguity and severe self-occlusion~\citep{gan2025handjoke}. 
To overcome these limitations, methods have shifted to parametric mesh recovery with strong anatomical priors (e.g., MANO)~\citep{Romero2017MANO, Moon2020I2LMeshNet, Baek2019Pushing, moon2020interhand2, Ge20193D, Hampali2020HOnnotate}, to constrain the shape and topology of reconstructed hands. However, anatomical plausibility does not by itself ensure accurate 3D localization. Such priors restrict the space of possible hand configurations but cannot replace explicit geometric constraints from image observations, particularly when depth cues are ambiguous or joints are occluded. This motivates exploiting calibrated multi-view observations to directly constrain joint positions and depth, rather than relying solely on prior-driven model fitting.

To overcome monocular bottlenecks, recent stereo frameworks like HandMvNet~\citep{Ali2025HandMvNet} and POEM~\citep{yang2023poem} utilize cross-view geometric triangulation for robust keypoint localization by resolving occlusions across viewpoints. Furthermore, precise kinematic features provide discriminative spatial cues that boost the performance of high-level gesture recognition~\citep{mucha2024my, Garcia-Hernando2018FPHA, Kwon2021H2O, zhao2023spatial}, motivating the use of pose trajectories as an input to gesture recognition.

\subsection{Event-based Hand Perception}
\label{subsec:event_hand_perception}

Event cameras offer high temporal resolution and a wide dynamic range, providing a complementary sensing modality for hand perception under rapid motion and challenging illumination~\citep{Gallego2022Event}.
While third-person event-based frameworks ~\citep{Amir2017low,Rudnev2021EventHands,Millerdurai2024Ev2Hands,Jiang2024EvHandPose,Jiang2024Complementing} have achieved high-fidelity perception, their third-person settings do not directly address the rapid ego-motion and perspective changes encountered in head-mounted setups.

Egocentric event-based perception additionally requires separating hand motion from background activity induced by camera motion. While early works addressed gesture recognition~\citep{Bhattacharyya2024Helios, Wang2025EgoEvGesture, Plizzari2022E2} or full-body motion~\citep{millerdurai2024eventego3d}, they generally lack fine-grained bimanual 3D articulations. Recent attempts at egocentric 3D hand mesh reconstruction rely on synthetic data~\citep{Hara2025EventEgoHands}, limiting real-world robustness. A critical gap persists: existing approaches either address high-level gesture semantics without providing kinematic pose information or focus on 3D pose estimation within monocular settings that suffer from inherent depth ambiguity and extreme scale variations. These limitations motivate real-world stereo pose annotations and frameworks that support both 3D pose estimation and gesture recognition under egocentric conditions.

\subsection{Stereo Fusion and Joint Perception Paradigms}
\label{subsec:stereo_fusion}
Stereo vision provides a principled physical solution to resolve the depth uncertainty inherent in monocular observations. In terms of 3D representations, dense volumetric lifting~\citep{philion2020lift, yang2023poem, ghasemzadeh2024mpl} constructs explicit voxel grids to enforce spatial consistency; however, dense 3D grids introduce substantial computation for a task whose outputs are sparse joint locations. Conversely, lightweight two-stage pipelines~\citep{Xu2023EVTIFNet, Iskakov2019Learnable} rely on analytical triangulation after 2D detection, making them highly vulnerable to occlusion-induced errors that propagate during lifting. Recently, Bird’s-Eye-View (BEV) projection~\citep{Li2022BEVFormer, Huang2021LSS} has emerged to bridge this gap, preserving 3D topological consistency with 2D-like efficiency. Yet, contemporary multi-view pose estimation frameworks~\citep{Ali2025HandMvNet, EgoPoseFormer2024} primarily employ implicit feature-level cross-attention rather than centering every refinement step on the projection of an explicit 3D joint hypothesis.
Furthermore, a critical architectural misalignment exists: standard BEV paradigms are tailored for dense scene occupancy, whereas hand pose estimation is fundamentally a sparse keypoint localization problem. Current egocentric stereo approaches~\citep{RealTimeMultiView2025, TwoViewpoints2024} frequently revert to dense volumes, lacking a dedicated mechanism to fuse sparse keypoint features under explicit geometric guidance. These approaches motivate a sparse fusion mechanism that explicitly links 3D joint hypotheses to their corresponding observations in multiple views.

Our approach maintains sparse 3D joint hypotheses and uses calibrated projection to retrieve joint-specific visual evidence. Global stereo feature memory supplies context without requiring dense 3D voxelization. This separates the role of geometric anchoring from that of feature fusion.

Table~\ref{tab:datasets} compares EgoEVHands with representative frame-based and event-based resources, including EgoEvGesture. EgoEvGesture provides the underlying egocentric stereo recordings and gesture labels; EgoEVHands advances this resource with dense 2D/3D hand annotations, cross-view supervision, segmentation masks, and a pose-oriented evaluation protocol. Together, these additions establish the first large-scale real-world stereo event dataset for joint egocentric 3D bimanual hand pose estimation and gesture recognition, supporting geometric and semantic hand perception within a unified dataset.

\begin{table}[pos=!htbp,width=\linewidth,abovecap=3pt,belowcap=3pt]
\centering
\caption{Comparison of hand-perception datasets. EgoEvGesture is included to distinguish the original gesture dataset from the pose-annotated EgoEVHands extension.}
\label{tab:datasets}

\begingroup
\fontsize{8}{9.6}\selectfont
\setlength{\tabcolsep}{2.4pt}
\renewcommand{\arraystretch}{1.08}
\setlength{\extrarowheight}{0pt}
\setlength{\aboverulesep}{2pt}
\setlength{\belowrulesep}{2pt}

\begin{tabular*}{\linewidth}{@{\extracolsep{\fill}}l*{10}{c}@{}}
\toprule
\textbf{Dataset} & \textbf{Modality} & \textbf{Volume} & \textbf{Classes} & \textbf{Gesture} & \textbf{2D} & \textbf{3D} & \textbf{2H} & \textbf{Real} & \textbf{1P} & \textbf{Res.} \\
\midrule
EgoGesture~\citep{Zhang2018EgoGesture} & RGB & 3M & 83 & $\checkmark$ & -- & -- & $\checkmark$ & $\checkmark$ & $\checkmark$ & 640 \\
FPHA~\citep{Garcia-Hernando2018FPHA} & RGB & 0.1M & 45 & $\checkmark$ & P & $\checkmark$ & -- & $\checkmark$ & $\checkmark$ & 1080 \\
EPIC-K~\citep{Damen2020EPIC} & RGB & 11M & 149 & $\checkmark$ & -- & -- & $\checkmark$ & $\checkmark$ & $\checkmark$ & 1080 \\
H20~\citep{Kwon2021H2O} & RGB & 0.6M & 36 & $\checkmark$ & $\checkmark$ & $\checkmark$ & $\checkmark$ & $\checkmark$ & $\checkmark$ & 720 \\
Asm101~\citep{Sener2022Assembly101} & RGB & 110M & 1k & $\checkmark$ & -- & $\checkmark$ & -- & $\checkmark$ & $\checkmark$ & 1080 \\
ARCTIC~\citep{Fan2023ARCTIC} & RGB & 2M & 11 & $\checkmark$ & -- & $\checkmark$ & $\checkmark$ & $\checkmark$ & $\checkmark$ & 2800 \\
HOI4D~\citep{Liu2022HOI4D} & RGB & 2M & 54 & $\checkmark$ & $\checkmark$ & $\checkmark$ & -- & $\checkmark$ & $\checkmark$ & 8000 \\
DVS128~\citep{Amir2017low} & Event & 1kS & 10 & $\checkmark$ & -- & -- & -- & $\checkmark$ & -- & 128 \\
N-EPIC~\citep{Plizzari2022E2} & Event & -- & 8 & $\checkmark$ & -- & -- & $\checkmark$ & -- & $\checkmark$ & 225 \\
EHoA~\citep{Chen2024Ehoa} & Event+RGB & 2kS & 8 & $\checkmark$ & -- & -- & -- & $\checkmark$ & -- & 346 \\
EvRealH~\citep{Jiang2024EvHandPose} & Event+RGB & 1.3h & -- & -- & -- & $\checkmark$ & -- & $\checkmark$ & -- & 720 \\
Ev2H-S~\citep{Millerdurai2024Ev2Hands} & Event & * & -- & -- & $\checkmark$ & $\checkmark$ & $\checkmark$ & -- & -- & 512 \\
Ev2H-R~\citep{Millerdurai2024Ev2Hands} & Event+RGB & 0.3h & -- & -- & -- & P & $\checkmark$ & $\checkmark$ & -- & 346 \\
EvHands~\citep{Rudnev2021EventHands} & Event & 100h & -- & -- & -- & $\checkmark$ & -- & -- & -- & 240 \\
EvEgo3D~\citep{millerdurai2024eventego3d} & Event & 2.1h & -- & -- & -- & $\checkmark$ & -- & B & $\checkmark$ & 320 \\
Helios~\citep{Bhattacharyya2024Helios} & Event & 600S & 7 & $\checkmark$ & -- & P & -- & $\checkmark$ & $\checkmark$ & 320 \\
EvRealH2~\citep{Jiang2024Complementing} & Event+RGB & 1.2h & -- & -- & -- & $\checkmark$ & -- & $\checkmark$ & -- & 346 \\
EgoEvGesture~\citep{Wang2025EgoEvGesture} & Event & 5kS & 38 & $\checkmark$ & -- & -- & $\checkmark$ & $\checkmark$ & $\checkmark$ & 720 \\
\midrule
EgoEVHands & Event & 2,498 pairs & 38 & $\checkmark$ & $\checkmark$ & $\checkmark$ & $\checkmark$ & $\checkmark$ & $\checkmark$ & 720 \\
\bottomrule
\end{tabular*}

\par\vspace{3pt}
\begin{minipage}{\linewidth}
\fontsize{7.5}{9}\selectfont\raggedright
$\checkmark$ = yes, -- = no, P = partial, and B = both.
Gesture indicates gesture/action labels; 2H denotes bimanual data;
Real denotes real-world acquisition; 1P denotes first-person observations;
and Res. denotes resolution in pixels.
Partial resources may enable joint derivation from RGB+D, meshes, or partial
annotations, but do not explicitly provide complete 2D joint ground truth.
Masks or action labels alone are not counted as 2D keypoint annotations;
2D keypoints are not assumed derivable from published 3D annotations without
explicit camera intrinsics/parameters or a documented
3D$\rightarrow$2D projection annotation pipeline.
Volume units are M = $10^{6}$ frames, k = $10^{3}$, S = samples,
m = minutes, h = hours, and * = $3.1\times10^{8}$ events.
EgoEVHands is counted in annotated stereo sequence pairs.
\end{minipage}

\endgroup
\end{table}

\section{EgoEVHands: A Large-Scale Pose-Annotated Stereo Event Dataset}
\label{sec:dataset}
\subsection{Dataset Acquisition and Composition}
\label{sec:dataset_acquisition}

EgoEVHands establishes the first large-scale real-world stereo event dataset for egocentric 3D bimanual hand pose estimation and gesture recognition. Built upon the stereo event recordings originally collected for EgoEvGesture~\citep{Wang2025EgoEvGesture}, it extends the original gesture-level labels with dense pose annotations for both single-hand and bimanual interactions. The dataset contains 2,498 annotated stereo sequence pairs from $10$ participants across $38$ gesture categories. Each pair comprises synchronized left- and right-view event sequences of the same gesture execution.

\begin{figure}[pos=!htbp,width=\linewidth,align=\centering,abovecap=4pt,belowcap=2pt]
\begin{minipage}[t]{0.48\linewidth}
\vspace{0pt}\centering
\includegraphics[width=\linewidth]{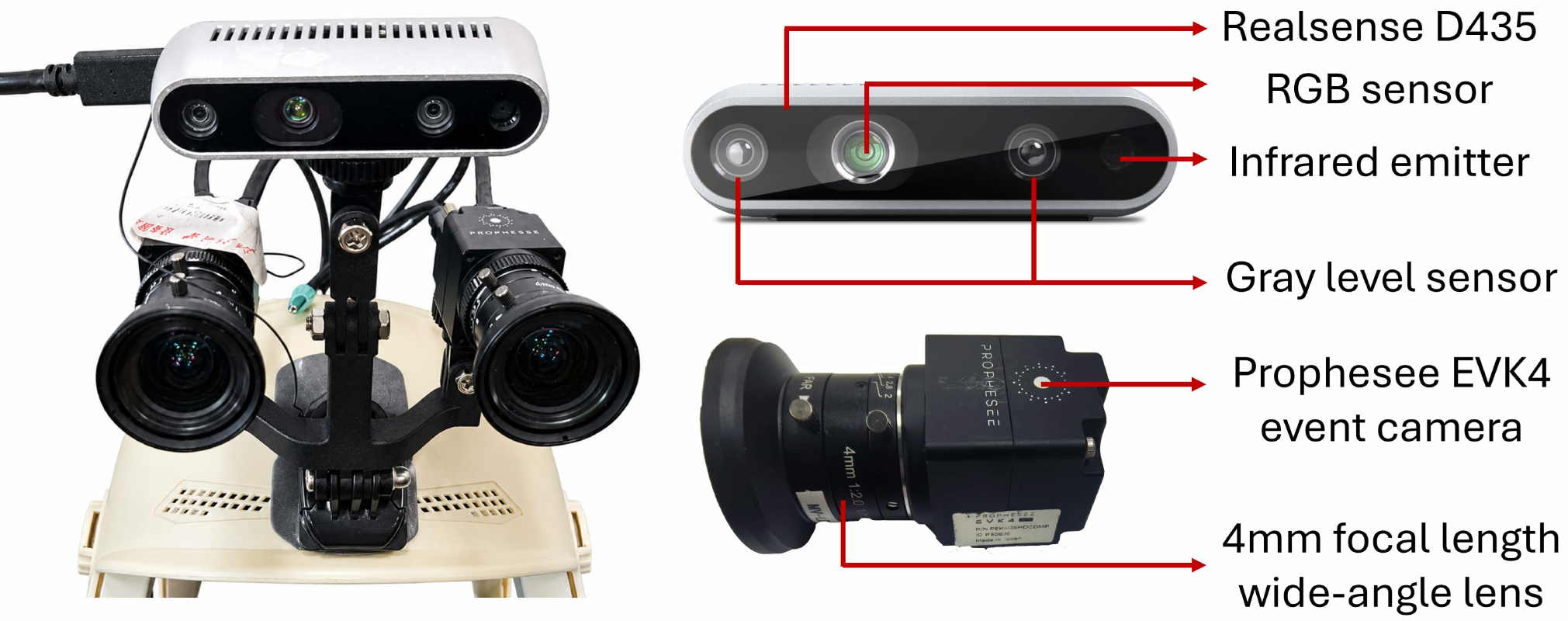}
\LocalFigureWidth
\caption{The head-mounted capture system employed for EgoEVHands data acquisition. The system integrates a RealSense D435 RGB-D camera and two Prophesee EVK4 event cameras, each equipped with a $4$\,mm focal length wide-angle lens. The cameras are rigidly mounted for synchronized egocentric recording.}
\label{fig:hmcs}
\end{minipage}\hfill
\begin{minipage}[t]{0.48\linewidth}
\vspace{0pt}\centering
\nextfloat
\begin{subfigure}[t]{0.48\linewidth}
\centering
\includegraphics[width=\linewidth]{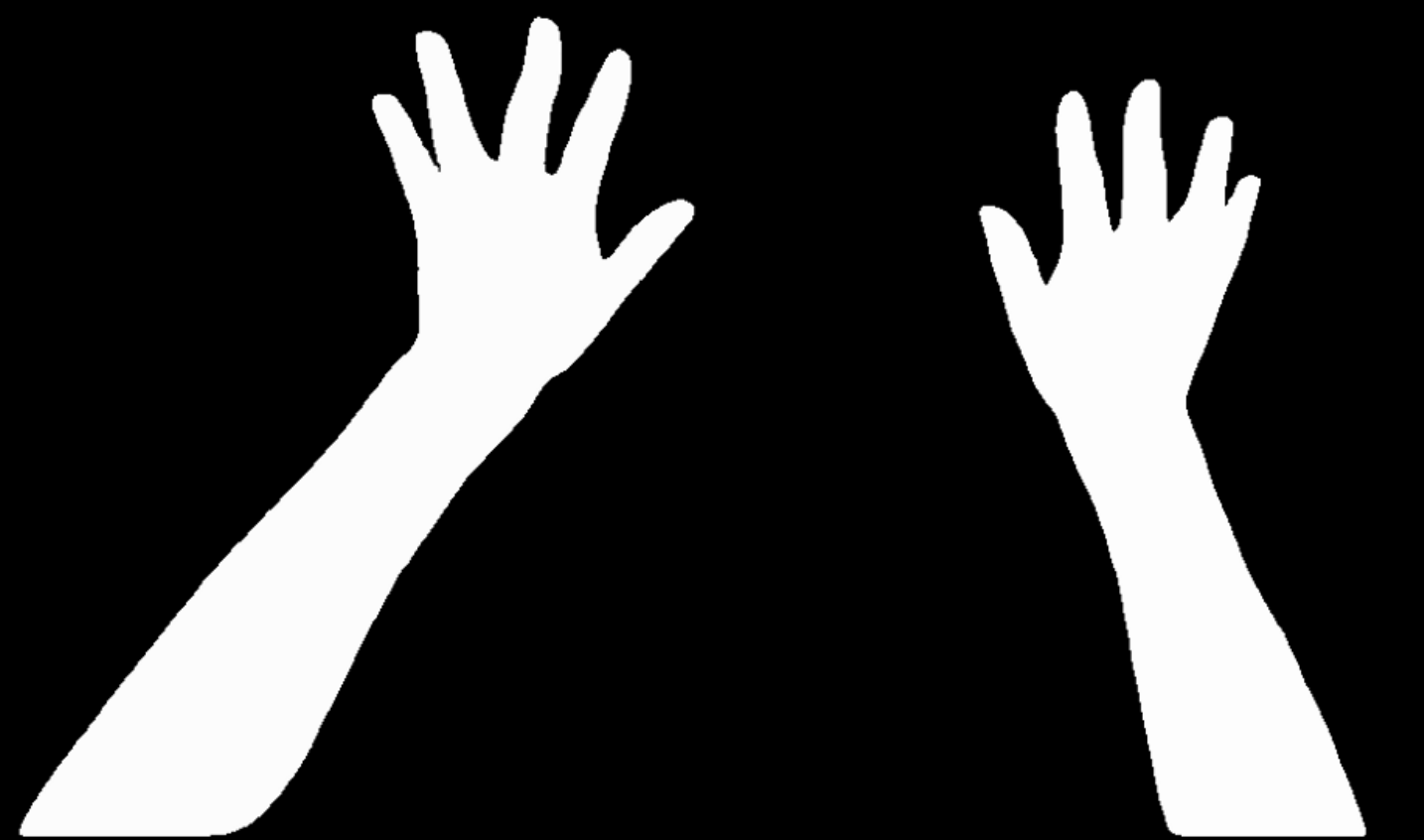}
\caption{Hand segmentation mask generated by Grounded-SAM.}
\label{fig:annotation:mask}
\end{subfigure}\hfill
\begin{subfigure}[t]{0.48\linewidth}
\centering
\includegraphics[width=\linewidth]{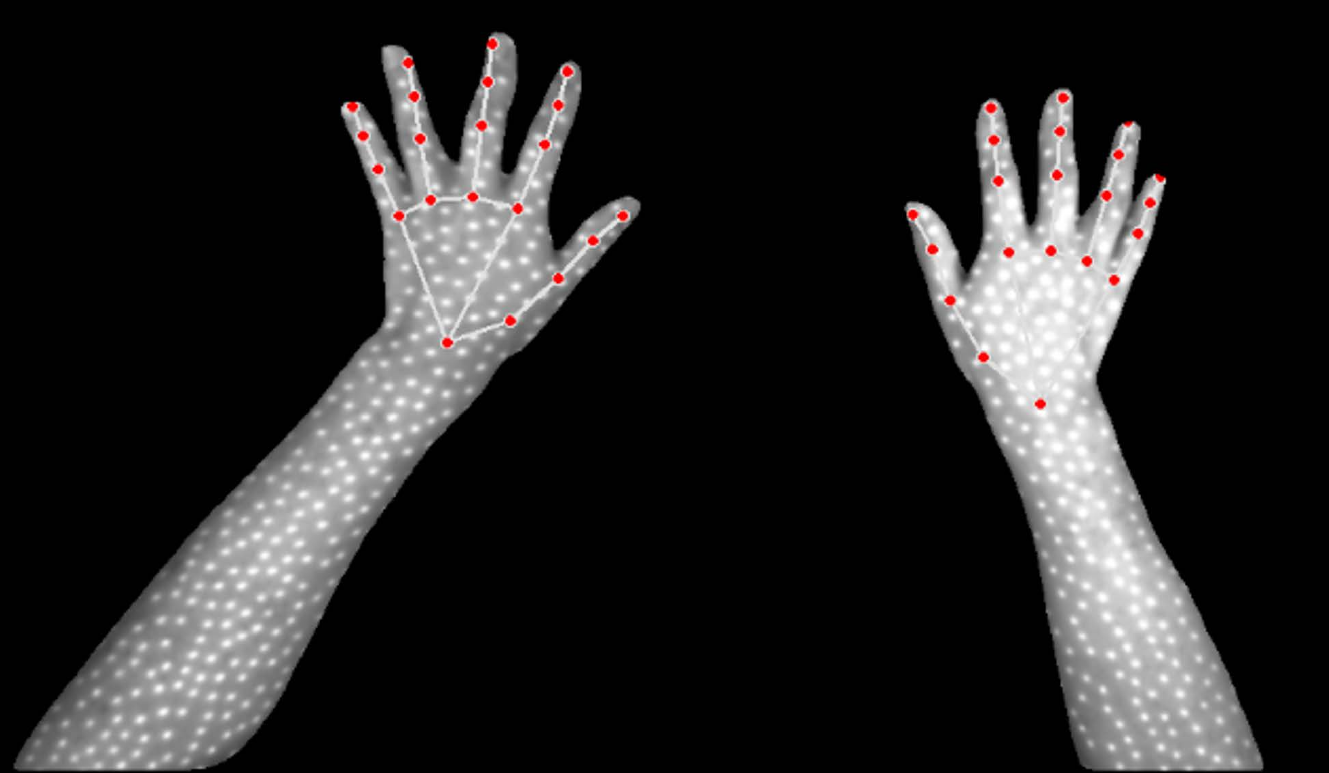}
\caption{Original 2D hand keypoints detected on the RealSense infrared image.}
\label{fig:annotation:2d}
\end{subfigure}
\LocalFigureWidth
\caption{Illustration of the intermediate annotation results in the proposed pipeline, including the automatically detected hand segmentation mask and the corresponding 2D hand keypoints.}
\end{minipage}
\end{figure}

The head-mounted acquisition system integrates two Prophesee EVK4 event cameras and an Intel RealSense D435 RGB-D camera, as shown in Fig.~\ref{fig:hmcs}. Each event camera has a resolution of $1280\times720$ and is equipped with a $4$\,mm focal length wide-angle lens. The cameras are rigidly mounted for synchronized egocentric recording. The acquisition setup and gesture protocol follow EgoEvGesture, while the present work uses RealSense depth observations to construct pose-level supervision. Recordings cover both normal and low-light conditions, providing real-world data for studying hand perception across different illumination levels.

Each annotated sample provides synchronized stereo event streams, corresponding RealSense RGB, infrared and depth observations, dense 2D/3D hand keypoints, and hand segmentation masks. The pose annotations additionally include projections into both event views and joint-level visibility and validity information. These complementary annotations support stereo geometric reconstruction, hand localization, and gesture recognition within a unified dataset. RealSense observations, Grounded-SAM, and MediaPipe are used for offline annotation; inference with EgoEV-HandPose requires only the stereo event streams and calibrated camera parameters.

\subsection{Annotation Pipeline}
\label{sec:annotation_pipeline}
The semi-automatic pipeline comprises 2D keypoint initialization and manual refinement, depth-based 3D reconstruction, temporal densification, and cross-view projection.

\textbf{2D initialization and refinement.}
Grounded-SAM~\citep{ren2024grounded} produces hand segmentation masks from the RealSense infrared images (Fig.~\ref{fig:annotation:mask}). MediaPipe~\citep{lugaresi2019mediapipe} detects $21$ keypoints per hand, or $42$ for two hands, on three image variants: the original, masked, and background-removed infrared images. The most reliable detections are selected, manually screened, and corrected when necessary to obtain validated 2D annotations (Fig.~\ref{fig:annotation:2d}).

\textbf{Depth-based reconstruction.}
For a validated joint at image coordinate $(u_j,v_j)$ with depth $d_j$, its position in the RealSense camera coordinate system is
\begin{equation}
\mathbf{p}^{\mathrm{RS}}_j=d_j\mathbf{K}_{\mathrm{RS}}^{-1}
\begin{bmatrix}u_j&v_j&1\end{bmatrix}^{\!\top},
\label{eq:annotation_backprojection}
\end{equation}
where $\mathbf{K}_{\mathrm{RS}}$ is the calibrated intrinsic matrix. Valid neighboring depth measurements are used when the joint depth is missing. Extrinsic transformations subsequently express the reconstructed points in the common coordinate system used for stereo projection.

\textbf{Temporal densification and validity.}
Linear interpolation in 3D is applied between adjacent validated annotations only when the temporal gap is no larger than five frames. Larger gaps remain invalid. Each keypoint has an annotation weight $w$: $w=1$ for a directly validated point, $0<w<1$ for an interpolated point, and $w=0$ for an invalid point. These weights and the associated status flags distinguish observed supervision from temporal interpolation; interpolated points are not described as directly measured annotations.

\textbf{Cross-view projection.}
The reconstructed 3D keypoints are transformed into the left and right event-camera coordinate systems using the calibrated extrinsics and projected into their image planes. Thus, event-view 2D keypoints are obtained from the 3D annotations, rather than from separate keypoint detections on the event images. Calibrated transformations also transfer the hand masks into the event views. The resulting annotations are stored together with joint-level visibility and validity information.

\subsection{Dataset Diversity}
\label{sec:dataset_diversity}

\textbf{Hand configurations and interaction complexity.}
The $38$ gesture categories span diverse hand configurations and motion patterns performed by $10$ participants. To characterize interaction complexity, the categories are organized into four representative groups: single-hand interaction, bimanual interaction without occlusion, self-occlusion, and mutual occlusion. This coverage includes both isolated hand movements and interacting hands with partially hidden joints, supporting the study of 3D hand reconstruction and gesture understanding under different occlusion conditions.

\textbf{Normal-light and low-light coverage.}
An important feature of EgoEVHands is the inclusion of real recordings under both normal-light and low-light conditions. Together with the range of hand interactions, this illumination coverage enables pose estimation and gesture recognition to be evaluated beyond a single lighting condition. The dataset supports scenario-wise comparisons between normal and low illumination, as well as analyses of performance across different interaction and occlusion categories. It therefore provides data for developing and evaluating illumination-robust egocentric hand perception.

\begin{figure}[pos=!htb,width=\linewidth,align=\centering,abovecap=0pt,belowcap=0pt]
\begin{minipage}[c]{.54\linewidth}
\centering
\includegraphics[width=\linewidth]{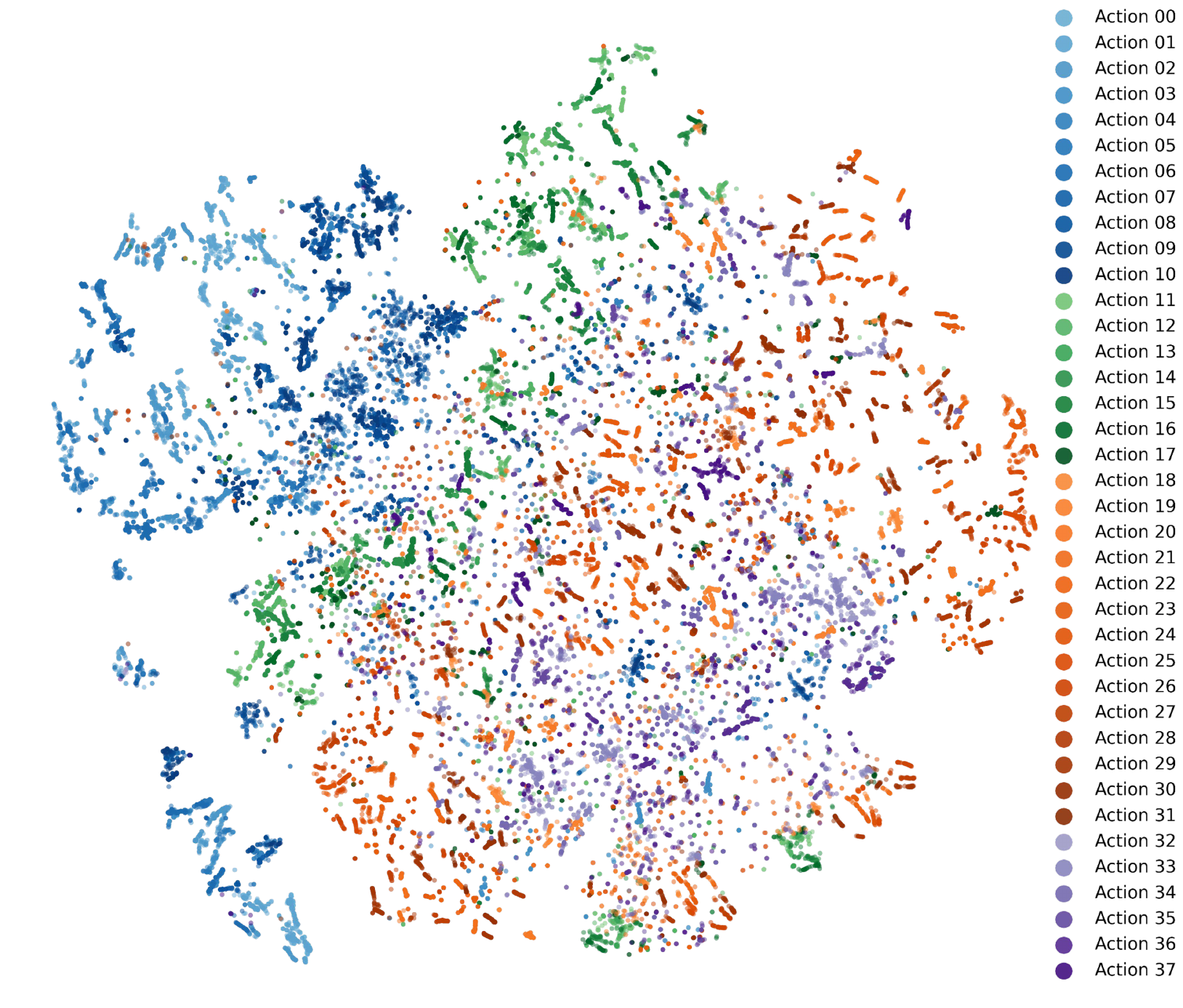}
\end{minipage}\hfill
\begin{minipage}[c]{.43\linewidth}
\LocalFigureWidth
\caption{t-SNE visualization of EgoEVHands. The embedding illustrates the distribution and diversity of interaction patterns across the $38$ gesture categories, covering single-hand interaction, bimanual interaction without occlusion, self-occlusion, and mutual occlusion.}
\label{fig:tsne}
\end{minipage}
\end{figure}

\textbf{Visualization of interaction diversity.}
Figure~\ref{fig:tsne} presents a t-SNE visualization of the dataset. Following the stereo organization of the recordings, the sequences are unfolded into frame-level samples, and the left and right views are treated independently. The embedding visualizes the distribution of the $38$ gesture categories and their four interaction groups, complementing the description of single-hand, bimanual, and occluded hand configurations. Together with the normal-light and low-light recordings, these interaction patterns reflect the diversity of the real-world observations provided by EgoEVHands.

\subsection{Data Release and Reproducibility}
\label{sec:data_release}

The planned release will include stereo event streams and timestamps, the RealSense infrared and depth observations used for annotation, dense 2D/3D hand keypoints, event-view projections, hand segmentation masks, annotation weights and validity flags, calibration parameters, gesture labels, and official data splits. Preprocessing and annotation-conversion scripts will accompany the data. The coordinate systems, joint ordering, timestamp convention, and annotation-weight definition will be documented to support reproducible use of the dataset for stereo hand pose estimation and gesture recognition.

\section{Methodology}
\label{sec:method}

\subsection{Problem Formulation and Framework Overview}
\label{sec:problem_overview}
Given synchronized stereo event streams, EgoEV-HandPose estimates the 3D positions of $J=42$ hand joints and predicts one of $38$ gesture labels. The left and right LNES inputs are processed by EgoBlaze with shared weights. For each view $v\in\{l,r\}$, it produces 2D coordinates $\hat{\mathbf{U}}^v\in\mathbb{R}^{J\times2}$ and intermediate visual features. Analytical stereo triangulation yields $\mathbf{P}^{(0)}\in\mathbb{R}^{J\times3}$. KeypointBEV performs exactly three learned updates in both training and inference:
\begin{equation}
(\mathcal{E}^{l},\mathcal{E}^{r})\longrightarrow
(\hat{\mathbf{U}}^{l},\hat{\mathbf{U}}^{r})\longrightarrow
\mathbf{P}^{(0)}\longrightarrow\mathbf{P}^{(1)}
\longrightarrow\mathbf{P}^{(2)}\longrightarrow\mathbf{P}^{(3)}.
\label{eq:overall_pose}
\end{equation}
The analytical initialization is distinct from every learned update. Throughout the method, $\mathbf{P}^{(k)}$ denotes the pose after $k$ learned updates, with $k=0$ reserved for initialization.

KeypointBEV adopts a BEV-inspired formulation to learn and refine hand-joint representations in a common 3D coordinate space while maintaining sparse 3D joint hypotheses rather than a dense voxelized BEV volume. We instantiate this formulation with spatially anchored joint queries, using each current 3D joint hypothesis to retrieve and aggregate features from both views. The resulting joint-centered representation combines local stereo evidence with global visual context for iterative 3D pose refinement. The final pose $\mathbf{P}^{(3)}$ is obtained and reprojected into both image planes to provide refined 2D trajectories. After normalization, the two views contribute $2\times42\times2=168$ coordinates per frame to a temporal Transformer to obtain the gesture.

\begin{figure}[pos=!htbp,width=\linewidth,align=\centering,abovecap=4pt,belowcap=2pt]
\includegraphics[width=\linewidth,height=.54\textheight,keepaspectratio]{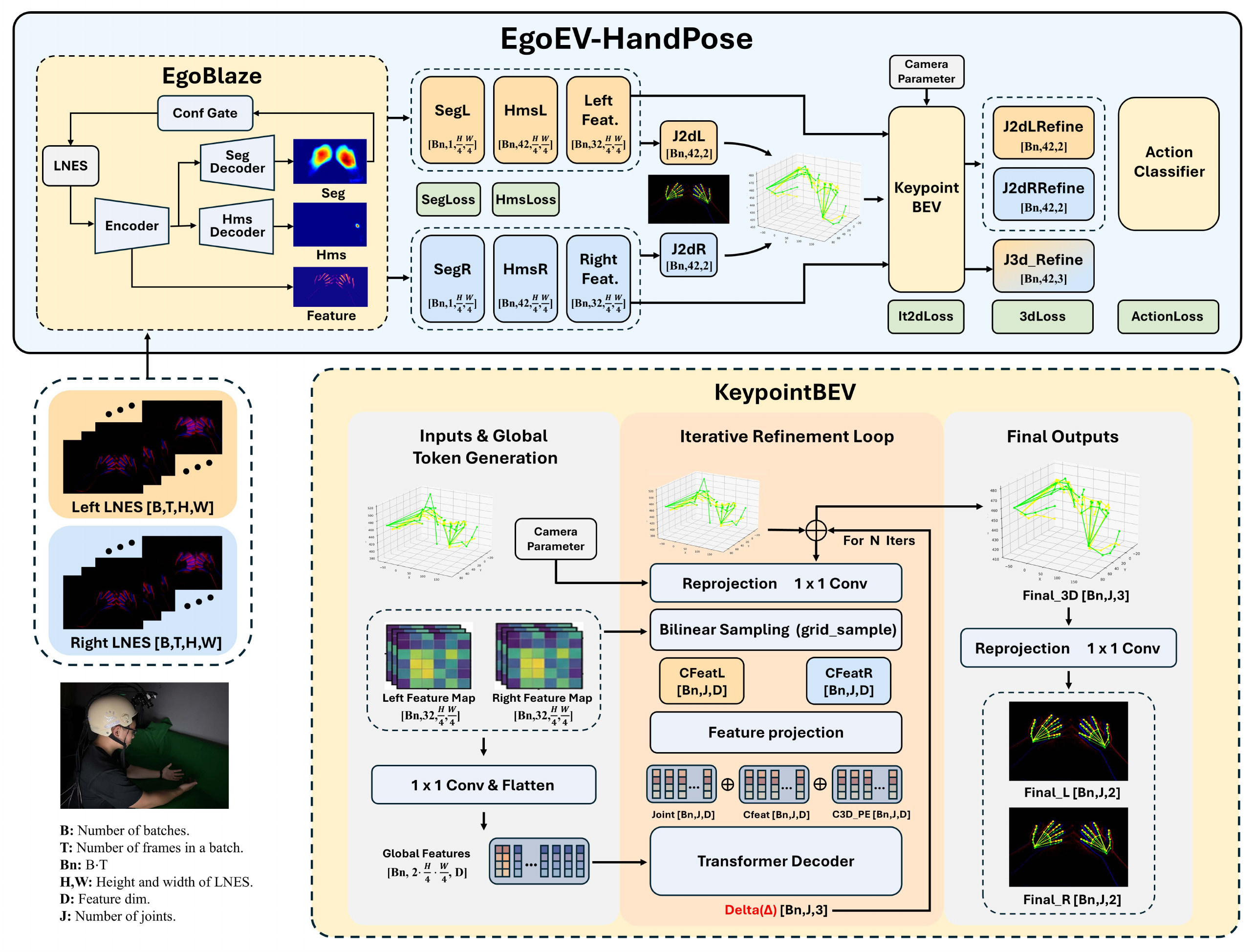}
  \caption{Overall pipeline of EgoEV-HandPose. Stereo events are represented as LNES and processed by a weight-sharing EgoBlaze backbone to obtain 2D heatmaps, segmentation masks, and local features. KeypointBEV refines the analytical stereo initialization through three learned updates. The final 3D pose is reprojected into both views, and the resulting 2D coordinate sequences are used for gesture recognition.}
  \label{fig:pipeline}
\end{figure}

\subsection{Stereo Event-based 2D Keypoint Estimation}
\label{sec:2d_hpe}
For each view, events within a temporal window $\Delta T$ are represented as Locally-Normalized Event Surfaces (LNES)~\citep{Rudnev2021EventHands}, denoted by $\mathbf{S}^v_t\in\mathbb{R}^{H\times W\times2}$. The left and right inputs are processed by the same EgoBlaze network~\citep{Wang2025EgoEvGesture}:
\begin{equation}
(\mathbf{H}^v_t,\mathbf{A}^v_t,\mathbf{F}^v_t)
=\operatorname{EgoBlaze}_{\theta}(\mathbf{S}^v_t),
\qquad v\in\{l,r\},
\label{eq:shared_egoblaze}
\end{equation}
where $\mathbf{H}^v_t$, $\mathbf{A}^v_t$, and $\mathbf{F}^v_t$ are the keypoint heatmaps, auxiliary hand segmentation output, and intermediate features. The common parameter set $\theta$ is shared across views; their event buffers and activations remain view-specific.

Internally, EgoBlaze incorporates EROS-based temporal event buffering, a Blaze-style convolutional encoder, and a Mamba bottleneck. These are components of the shared EgoBlaze implementation, not alternative backbones or a replacement for its LNES representation. The encoder channel dimensions are $16$, $32$, $64$, $128$, and $192$; the bottleneck contains two Mamba layers with hidden dimension $192$. The internal Confidence Gate~\citep{Millerdurai2024Ev2Hands} suppresses background responses before the heatmap and segmentation decoding branches. This feature-gating mechanism does not append confidence values to the gesture-classifier input.

For an input of $720\times1280$ pixels, the heatmap decoder predicts $\mathbf{H}^{v}\in\mathbb{R}^{J\times180\times320}$. A spatial softmax is applied independently to each joint heatmap, and the image coordinate is recovered by a differentiable expectation:
\begin{equation}
\hat{\mathbf{u}}^v_j
=4\sum_{x,y}\operatorname{softmax}(\beta\mathbf{H}^v_j)_{x,y}
\begin{bmatrix}x&y\end{bmatrix}^{\!\top},
\qquad \beta=100.
\label{eq:softargmax}
\end{equation}
The factor of four converts heatmap coordinates into the original image coordinate system. Stacking the joint coordinates as rows gives $\hat{\mathbf{U}}^v$. These coordinates provide the observations for triangulation, while intermediate feature maps provide the visual evidence for refinement.

\subsection{Analytical Stereo 3D Initialization}
\label{sec:stereo_initialization}
The initialization contains no learned 3D refinement. Predicted image coordinates are corrected using the calibrated intrinsics and distortion parameters, with five radial-undistortion iterations. Each corrected point defines a ray in a common coordinate system,
\begin{equation}
\mathbf{r}^v_j(t)=\mathbf{c}^v+t\mathbf{d}^v_j,
\label{eq:camera_ray}
\end{equation}
where $\mathbf{c}^v$ is the camera center and $\mathbf{d}^v_j$ is the unit ray direction. Localization errors generally prevent the stereo rays from intersecting. Their closest points are found by solving
\begin{equation}
(t^{l*}_j,t^{r*}_j)=\arg\min_{t_l,t_r}
\left\|\mathbf{c}^{l}+t_l\mathbf{d}^{l}_j
-\mathbf{c}^{r}-t_r\mathbf{d}^{r}_j\right\|_2^2,
\label{eq:triangulation}
\end{equation}
and the midpoint defines the initial 3D joint:
\begin{equation}
\mathbf{p}^{(0)}_j=\tfrac12\left(
\mathbf{c}^{l}+t^{l*}_j\mathbf{d}^{l}_j+
\mathbf{c}^{r}+t^{r*}_j\mathbf{d}^{r}_j\right).
\label{eq:pinit}
\end{equation}
The resulting points form $\mathbf{P}^{(0)}$. Validity masks handle degenerate configurations, non-positive-depth solutions, invisible joints, and non-finite estimates during training.

\subsection{KeypointBEV: Sparse Reprojection-guided 3D Refinement}
\label{sec:keypoint_bev}
\label{sec:keypointbev}
Stereo initialization can be sensitive to image localization errors and occlusions. KeypointBEV therefore retains an explicit 3D hypothesis for every joint and repeatedly retrieves stereo evidence at the locations implied by that hypothesis.

\textbf{Global stereo feature memory.}
Each intermediate feature map $\mathbf{F}^v$ is spatially pooled and projected to $D=512$ channels using a $1\times1$ convolution. The flattened view-specific maps are concatenated to form the global memory:
\begin{equation}
\begin{aligned}
\mathbf{M}^v&=\operatorname{Conv}_{1\times1}(\operatorname{Pool}(\mathbf{F}^v)),\\
\mathbf{M}&=\operatorname{Concat}(\operatorname{Flatten}(\mathbf{M}^l),
\operatorname{Flatten}(\mathbf{M}^r))\in\mathbb{R}^{N_m\times D}.
\end{aligned}
\label{eq:stereo_memory}
\end{equation}
Here $N_m$ is the combined number of spatial tokens. This memory supplies stereo context without constructing a dense 3D grid.

\textbf{Sparse queries and geometric encoding.}
The module maintains $J$ joint queries $\mathbf{Q}^{(k)}\in\mathbb{R}^{J\times D}$. Each current joint position is normalized with respect to a predefined 3D coordinate range, positionally encoded, and modulated by a learnable sigmoid gate:
\begin{equation}
\mathbf{e}^{(k)}_j=\operatorname{PE}(\operatorname{Norm}(\mathbf{p}^{(k)}_j)),
\qquad
\tilde{\mathbf{e}}^{(k)}_j=\sigma(\mathbf{g})\odot\mathbf{e}^{(k)}_j.
\label{eq:spatial_enc}
\end{equation}
The gated encoding provides the query with its current geometric hypothesis. The normalization range and gate parameterization are implementation settings, not additional anatomical constraints.

\textbf{Reprojection-based visual evidence.}
Let $\Pi^v$ denote calibrated projection into view $v$. The projected image coordinate is transformed into the normalized sampling coordinates associated with the feature map:
\begin{equation}
\mathbf{u}^{v,(k)}_j=\Pi^v(\mathbf{p}^{(k)}_j),
\qquad
\hat{\mathbf{u}}^{v,(k)}_j=\mathcal{N}_v(\mathbf{u}^{v,(k)}_j).
\label{eq:norm_coord}
\end{equation}
The operator $\mathcal{N}_v$ accounts for image-to-feature scaling and the grid-sampling coordinate convention. 
Differentiable bilinear interpolation retrieves the local features:
\begin{equation}
\mathbf{h}^{v,(k)}_j=\mathcal{B}(\mathbf{F}^v,\hat{\mathbf{u}}^{v,(k)}_j),
\qquad
\mathbf{h}^{(k)}_j=\operatorname{Fuse}(\mathbf{h}^{l,(k)}_j,\mathbf{h}^{r,(k)}_j).
\label{eq:bilinear_sampling}
\end{equation}
Unlike unconditional concatenation of stereo feature maps, the sampled evidence depends on the current 3D hypothesis. The fused joint features form $\mathbf{H}^{(k)}\in\mathbb{R}^{J\times D}$.

\textbf{Transformer refinement and bounded updates.}
The geometric encodings and visual evidence are incorporated into the joint queries before cross-attention to the global memory:
\begin{equation}
\mathbf{Z}^{(k)}=\operatorname{TransformerDecoder}
\big(\mathbf{Q}^{(k)}+\tilde{\mathbf{E}}^{(k)}+\mathbf{H}^{(k)},\mathbf{M}\big).
\label{eq:transformer_decoder}
\end{equation}
The decoder has one layer, eight attention heads, and feed-forward dimension $2048$. A $512\rightarrow256\rightarrow3$ regression head maps each output token to $\Delta\mathbf{p}^{(k)}_j$. The update is
\begin{equation}
\mathbf{p}^{(k+1)}_j=\mathbf{p}^{(k)}_j+
\alpha^{(k)}\operatorname{clip}\big(s_{\Delta}\Delta\mathbf{p}^{(k)}_j,
-\tau^{(k)},\tau^{(k)}\big),
\label{eq:pose_update}
\end{equation}
where $s_{\Delta}$ is a learnable global scale, $\alpha^{(k)}$ is an iteration-specific update factor, and $\tau^{(k)}$ bounds the correction. The decoder and regression head are shared across the three updates, indexed by $k=0,1,2$. Both training and inference end at $\mathbf{P}^{(3)}$; the analytical initialization $\mathbf{P}^{(0)}$ is not counted as a learned refinement iteration.

\subsection{Training Objectives and End-to-end Optimization}
\label{sec:multi_task_optimization}
\label{sec:training_objectives}
Training uses staged initialization followed by joint end-to-end fine-tuning. The shared EgoBlaze estimator is first pretrained using heatmap mean-squared error and segmentation binary cross-entropy:
\begin{equation}
\mathcal{L}_{\mathrm{2D}}=\lambda_{\mathrm{hms}}\mathcal{L}_{\mathrm{Hms}}
+\lambda_{\mathrm{seg}}\mathcal{L}_{\mathrm{Seg}}.
\label{eq:2d_training}
\end{equation}
EgoBlaze is temporarily frozen during the initial KeypointBEV warm-up, and the temporal classifier is initialized using cached 2D keypoint sequences. These initialization stages do not define the final training regime: the complete network is subsequently unfrozen and jointly optimized, including classification gradients into the pose estimator.

\textbf{Three-step refinement supervision.}
The refinement depth is $N=3$ throughout training and inference. Let $\mathcal{V}$ be the set of valid joints and $\mathbf{p}^{*}_j$ the 3D target. The 3D loss and final-stage supervision are
\begin{equation}
\ell_{3D}^{(k)}=\frac{1}{|\mathcal{V}|}\sum_{j\in\mathcal{V}}
\operatorname{SmoothL1}_{\beta=10\,\mathrm{mm}}
(\mathbf{p}^{(k)}_j-\mathbf{p}^{*}_j),
\qquad
\mathcal{L}_{\mathrm{final3D}}=\ell_{3D}^{(3)}.
\label{eq:final3d_loss}
\end{equation}
Each refined 3D prediction is also projected into both views and compared with the 2D targets:
\begin{equation}
\ell_{2D}^{(k)}=\frac{1}{|\mathcal{V}|}
\sum_{j\in\mathcal{V}}\sum_{v\in\{l,r\}}
\left\|\Pi^v(\mathbf{p}^{(k)}_j)-\mathbf{u}^{v,*}_j\right\|_1.
\label{eq:aux2d_loss}
\end{equation}
The auxiliary objectives supervise all three reprojected outputs and the first two 3D outputs:
\begin{equation}
\mathcal{L}_{\mathrm{aux2D}}=\sum_{k=1}^{3}w^{2D}_k\ell_{2D}^{(k)},
\qquad
\mathcal{L}_{\mathrm{aux3D}}=\sum_{k=1}^{2}w^{3D}_k\ell_{3D}^{(k)}.
\label{eq:auxiliary_losses}
\end{equation}
The terminal 3D output is supervised separately, avoiding its duplication in the auxiliary 3D sum. The refinement objective is
\begin{equation}
\mathcal{L}_{\mathrm{refine}}=
0.5\,\mathcal{L}_{\mathrm{final3D}}
+0.02\,\mathcal{L}_{\mathrm{aux2D}}
+0.1\,\mathcal{L}_{\mathrm{aux3D}}.
\label{eq:total_refine_loss}
\end{equation}
The progressive stage weights $w^{2D}_k$ and $w^{3D}_k$ emphasize later predictions and are distinct from the annotation-source weights in Sec.~\ref{sec:annotation_pipeline}.

\textbf{Joint end-to-end fine-tuning.}
After initialization, online predictions replace cached keypoint inputs. The full objective combines heatmap, segmentation, 3D refinement, and gesture supervision:
\begin{equation}
\mathcal{L}_{\mathrm{joint}}
=\lambda_{\mathrm{hms}}\mathcal{L}_{\mathrm{Hms}}
+\lambda_{\mathrm{seg}}\mathcal{L}_{\mathrm{Seg}}
+\mathcal{L}_{\mathrm{refine}}
+\lambda_{\mathrm{act}}\mathcal{L}_{\mathrm{Act}},
\qquad
\mathcal{L}_{\mathrm{Act}}=\mathcal{L}_{\mathrm{CE}}(\hat{\mathbf{y}},y).
\label{eq:joint_objective}
\end{equation}
The heatmap, segmentation, and action weights are $\lambda_{\mathrm{hms}}=0.05$, $\lambda_{\mathrm{seg}}=1.0$, and $\lambda_{\mathrm{act}}=10.0$. Classification inputs are not detached: the action loss propagates through coordinate normalization and stereo reprojection into KeypointBEV and the shared EgoBlaze network. Thus, pose estimation and gesture recognition are jointly optimized rather than connected only at inference.

\subsection{Pose-based Gesture Recognition}
\label{sec:gesture_recognition}
The temporal classifier uses refined stereo 2D coordinate trajectories. Specifically, the final 3D pose is reprojected into both views as $\mathbf{U}^{v}_{\mathrm{ref},t}=\Pi^v(\mathbf{P}^{(3)}_t)$. For each view and hand, the coordinates are centered at the wrist and normalized by the wrist-to-middle-finger-MCP distance:
\begin{equation}
\bar{\mathbf{u}}^v_{j,t}=\frac{\mathbf{u}^v_{j,t}-\mathbf{u}^v_{\mathrm{wrist},t}}
{\|\mathbf{u}^v_{\mathrm{middle\text{-}MCP},t}-\mathbf{u}^v_{\mathrm{wrist},t}\|_2+\epsilon},
\qquad \epsilon=10^{-5},
\label{eq:action_normalization}
\end{equation}
where $\mathbf{u}^v_{j,t}$ is a refined 2D coordinate. Concatenating both views gives the coordinate-only temporal input:
\begin{equation}
\mathbf{x}_t=\operatorname{Concat}
\big(\operatorname{vec}(\bar{\mathbf{U}}^l_t),
\operatorname{vec}(\bar{\mathbf{U}}^r_t)\big)\in\mathbb{R}^{168},
\qquad
\mathbf{X}\in\mathbb{R}^{T\times168}.
\label{eq:action_input}
\end{equation}
The $168$ dimensions are exactly $2$ views $\times\,42$ joints $\times\,2$ coordinates; no confidence scores are concatenated. For the $48$-frame temporal window, $T=48$. A linear layer maps each frame from $168$ to $256$ dimensions. A learned classification token and sinusoidal positional encodings precede a three-layer Transformer encoder with eight attention heads, feed-forward dimension $1024$, and dropout $0.4$. The classification-token output is layer-normalized and passed through a $256\rightarrow512\rightarrow38$ MLP, followed by softmax, to obtain $\hat{\mathbf{y}}$.

Cached 2D sequences are used only to initialize the temporal head. During joint fine-tuning, its inputs are generated by the current pose network and remain differentiably connected to it. At inference, the same coordinate-only path is used without ground-truth poses or confidence inputs.

\section{Experiments}
\label{sec:experiments}

\subsection{Evaluation Metrics \& Implementation Details}
The proposed framework is evaluated using several quantitative metrics: 2D Error ($px$) for keypoint localization accuracy in the image plane; Mean Per-Joint Position Error (MPJPE) and Procrustes-aligned MPJPE (PA-MPJPE) in millimeters for absolute and structural 3D pose accuracy; the Area Under the Curve (AUC) of the Percentage of Correct Keypoints (PCK) calculated across thresholds from 0 to 50\,mm; and Action Accuracy (\%) for the $38$ interaction categories. 
We use two data-splitting protocols following \citet{Wang2025EgoEvGesture}: HETER holds out participants 1, 2, and 9 to evaluate cross-subject generalization; HOMO uses repetitions 0--2 for training and the remaining repetitions for testing within subjects.

The experiments are implemented in PyTorch on a single NVIDIA A800 GPU using Adam with an initial learning rate of $1.0\times10^{-4}$. Following staged initialization, joint end-to-end fine-tuning proceeds for $25$ epochs with a learning rate of $5.0\times10^{-5}$, decayed by $0.1$ every $10$ epochs, and gradient clipping at a maximum norm of $5.0$. Both views share EgoBlaze weights, KeypointBEV performs three updates in training and inference, and the temporal input is $\mathbf{X}\in\mathbb{R}^{T\times168}$ with $T=48$. The classification loss updates the pose network as described in Sec.~\ref{sec:training_objectives}.

\begin{table}[pos=!tb,width=\linewidth,abovecap=2pt,belowcap=2pt]
\caption{Comprehensive quantitative comparison on EgoEVHands. Overall accuracy,
scenario-wise error, and computational efficiency are aligned on the same row.
All error and complexity metrics are lower-is-better.}
\label{tab:main_results}
\centering
\fontsize{7.8}{9.5}\selectfont
\setlength{\tabcolsep}{1.8pt}
\renewcommand{\arraystretch}{1.12}
\begin{tabular*}{\linewidth}{@{\extracolsep{\fill}}lcc*{12}{r}@{}}
\toprule
\multirow{3}{*}{\textbf{Method}} & \multirow{3}{*}{\textbf{Venue}} & \multirow{3}{*}{\textbf{Mod.}} &
\multicolumn{3}{c}{\textbf{Overall accuracy}} & \multicolumn{7}{c}{\textbf{Scenario-wise MPJPE (mm)}} & \multicolumn{2}{c}{\textbf{Efficiency}}\\
\cmidrule(lr){4-6}\cmidrule(lr){7-13}\cmidrule(lr){14-15}
& & & \multirow{2}{*}{M-2D} & \multirow{2}{*}{M-3D} & \multirow{2}{*}{PA-M3D} &
\multirow{2}{*}{Norm.} & \multirow{2}{*}{Low} & \multirow{2}{*}{Sing.} &
\multicolumn{4}{c}{Bimanual} & \multirow{2}{*}{Params} & \multirow{2}{*}{GFLOPs}\\
\cmidrule(lr){10-13}
& & & & & & & & & All & Inter. & Self & Mut. & & \\
\midrule
\multicolumn{15}{@{}l}{\textbf{HOMO}\enspace\textit{(homogeneous distribution)}}\\[2pt]
HandMvNet\textsuperscript{a} & VISAPP'25 & F-B-1/3 & 29.64 & 36.43 & 26.97 & 34.81 & 37.20 & 36.09 & 38.12 & 37.49 & 37.82 & 39.21 & 85.9 & 87.7 \\
Ev2Hands\textsuperscript{b} & CVPR'24 & E-M-3 & 35.47 & 54.31 & 30.74 & 53.34 & 54.77 & 54.84 & 51.71 & 52.10 & 51.64 & 51.48 & 4.50 & 21.50 \\
EvHandPose\textsuperscript{c} & CVPR'24 & E-M-3 & 19.38 & 33.50 & 22.53 & 32.60 & 33.93 & 33.50 & -- & -- & -- & -- & 47.4 & \textbf{3.69} \\
EventEgo3D\textsuperscript{d} & CVPR'23 & E-M-1 & 27.24 & 36.61 & 27.55 & 37.23 & 36.32 & 36.28 & 38.24 & 36.90 & 100.1 & 38.22 & \textbf{1.38} & 5.53 \\
\rowcolor[HTML]{E8F4FF}
\textbf{Ours}\textsuperscript{*} & \textbf{--} & \textbf{E-B-1} & \textbf{11.48} & \textbf{22.03} & \textbf{18.96} & \textbf{22.31} & \textbf{21.90} & \textbf{22.04} & \textbf{21.99} & \textbf{21.65} & \textbf{22.09} & \textbf{22.13} & 5.88 & 19.86 \\
\midrule
\multicolumn{15}{@{}l}{\textbf{HETER}\enspace\textit{(heterogeneous distribution)}}\\[2pt]
HandMvNet\textsuperscript{a} & VISAPP'25 & F-B-1/3 & 38.41 & 71.87 & 34.83 & 62.16 & 82.43 & 68.31 & 72.62 & 69.98 & 73.97 & 73.64 & 85.9 & 87.7 \\
Ev2Hands\textsuperscript{b} & CVPR'24 & E-M-3 & 68.90 & 115.13 & 82.30 & 118.39 & 111.66 & 120.46 & 114.05 & 115.34 & 112.69 & 114.30 & 4.50 & 21.50 \\
EvHandPose\textsuperscript{c} & CVPR'24 & E-M-3 & 28.15 & 60.20 & 26.00 & 58.49 & 62.06 & 60.20 & -- & -- & -- & -- & 47.4 & \textbf{3.69} \\
EventEgo3D\textsuperscript{d} & CVPR'23 & E-M-1 & 42.10 & 72.76 & 36.50 & 70.49 & 75.21 & 61.51 & 75.05 & 67.34 & 73.06 & 84.35 & \textbf{1.38} & 5.53 \\
\addlinespace[2pt]
\multicolumn{15}{@{}l}{\textit{Lifting module analysis using our fixed 2D backbone}}\\[2pt]
EventEgo3D\textsuperscript{d} & CVPR'23 & E-M-1 & 16.64 & 51.27 & 23.40 & 48.31 & 54.47 & 45.29 & 52.49 & 48.67 & 51.63 & 56.97 & 2.31 & 12.00 \\
BEVFormer\textsuperscript{e} & ECCV'22 & F-B-3 & 16.64 & 32.03 & 22.85 & 31.58 & 32.52 & 34.41 & 31.55 & 31.12 & 31.35 & 32.16 & 18.14 & 374.14 \\
\rowcolor[HTML]{E8F4FF}
\textbf{Ours}\textsuperscript{*} & \textbf{--} & \textbf{E-B-1} & \textbf{16.64} & \textbf{30.54} & \textbf{21.08} & \textbf{31.49} & \textbf{29.50} & \textbf{33.88} & \textbf{29.86} & \textbf{27.43} & \textbf{30.89} & \textbf{31.02} & 5.88 & 19.86 \\
\bottomrule
\end{tabular*}
\par\smallskip
\begin{minipage}{\linewidth}
\fontsize{7.5}{9.2}\selectfont\raggedright
\textit{Metrics:} M-2D: mean pixel joint error (px); M-3D: mean per-joint position error (mm);
PA-M3D: Procrustes-aligned mean per-joint position error (mm); Params: parameters (M).
Norm./Low: normal/low light; Sing.: single-hand; All: all bimanual cases;
Inter.: interaction without occlusion; Self/Mut.: self/mutual occlusion.
\textsuperscript{*}Ours excludes the action classifier for comparison with pose-only methods.\par
\textit{Sources:}
\textsuperscript{a}HandMvNet~\citep{Ali2025HandMvNet};
\textsuperscript{b}Ev2Hands~\citep{Millerdurai2024Ev2Hands};
\textsuperscript{c}EvHandPose~\citep{Jiang2024EvHandPose};
\textsuperscript{d}EventEgo3D~\citep{millerdurai2024eventego3d};
\textsuperscript{e}BEVFormer~\citep{Li2022BEVFormer}.
\end{minipage}
\end{table}

\subsection{Main Results and Robustness Analysis}
\label{sec:main_results}

\textbf{Quantitative Evaluation across Distributions}: 
As detailed in Table~\ref{tab:main_results}, EgoEV-HandPose achieves state-of-the-art 3D pose estimation performance, with MPJPEs of $22.03$\,mm (HOMO) and $30.54$\,mm (HETER). The HOMO results demonstrate accurate reconstruction within subjects, while the HETER results establish strong cross-subject generalization to unseen participants and their gesture sequences. Together, these evaluations show that the proposed geometric formulation remains effective across both subject-consistent and subject-independent settings.

\textbf{Robustness to Lighting and Occlusion}: 
The framework demonstrates remarkable stability against environmental variations. In both HOMO and HETER distributions, the performance remains relatively invariant to lighting conditions; for instance, the error gap between normal and low-light scenarios is minimal (\textit{e.g.}, $31.49$\,mm \textit{vs.} $29.50$\,mm in the HETER distribution). 
Furthermore, our analysis of hand interaction complexity reveals that the model is particularly adept at handling bimanual interactions ($29.86$\,mm in the HETER distribution). 
Even in the presence of mutual occlusion (Mut.), which is inherently challenging for event cameras due to the high density of overlapping signals, our method achieves reliable localization ($22.13$\,mm in the HOMO distribution and $31.02$\,mm in the HETER distribution). 
These results demonstrate exceptional robustness under low illumination and severe self- and mutual occlusions. Complementary stereo observations provide spatial constraints for recovering partially occluded joints, while event-based features preserve informative hand-motion cues under challenging illumination. Their combination enables reliable geometric reconstruction during complex egocentric bimanual interactions.

\begin{figure}[pos=!htbp,width=\linewidth,align=\centering,abovecap=4pt,belowcap=2pt]
\begin{subfigure}[t]{.24\linewidth}
\centering
\includegraphics[width=\linewidth]{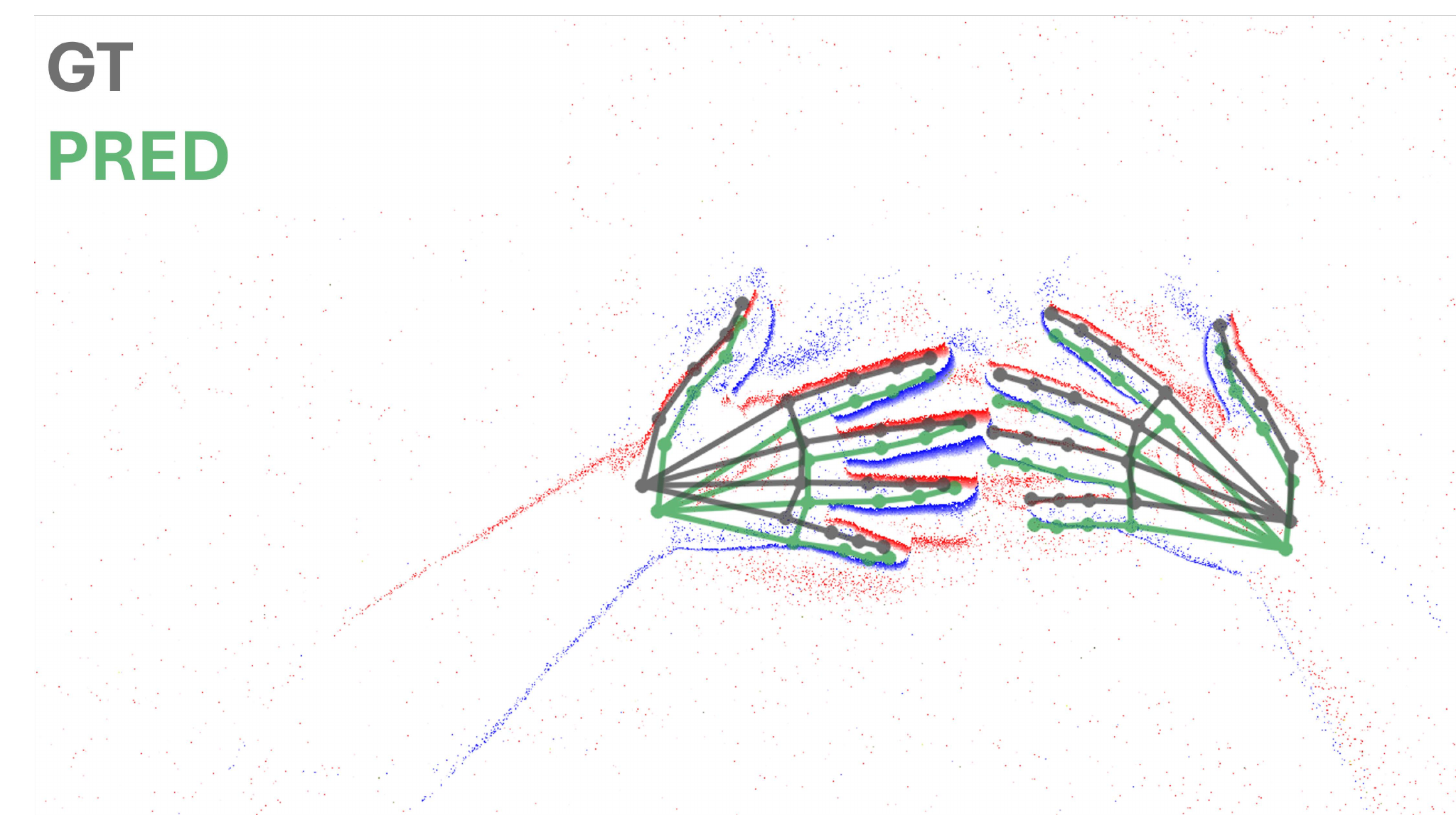}
\caption{Normal light scenarios.}
\label{fig:sub_normal}
\end{subfigure}
\hfill
\begin{subfigure}[t]{.24\linewidth}
\centering
\includegraphics[width=\linewidth]{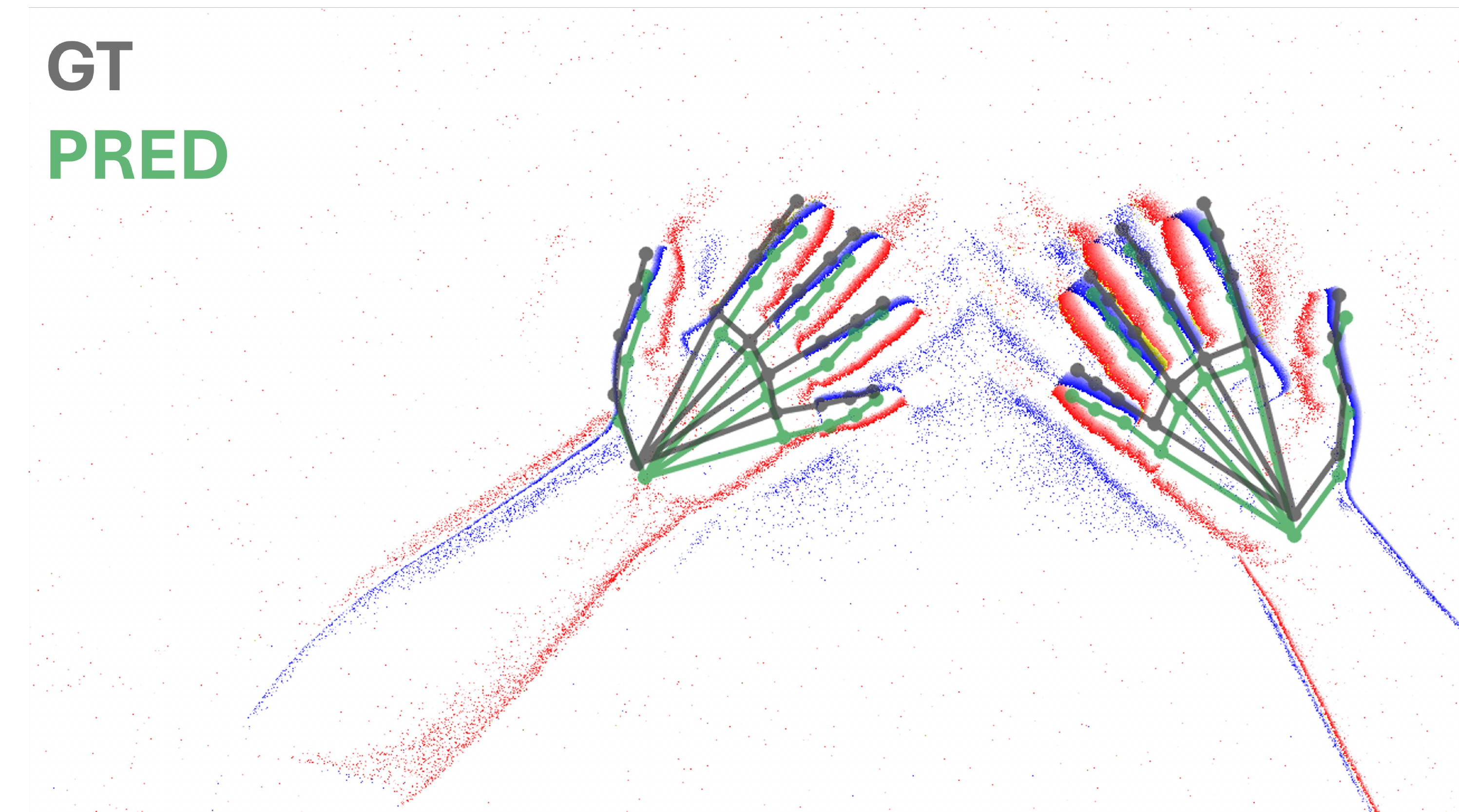}
\caption{Low-light scenarios.}
\label{fig:sub_pred}
\end{subfigure}
\hfill
\begin{subfigure}[t]{.24\linewidth}
\centering
\includegraphics[width=\linewidth]{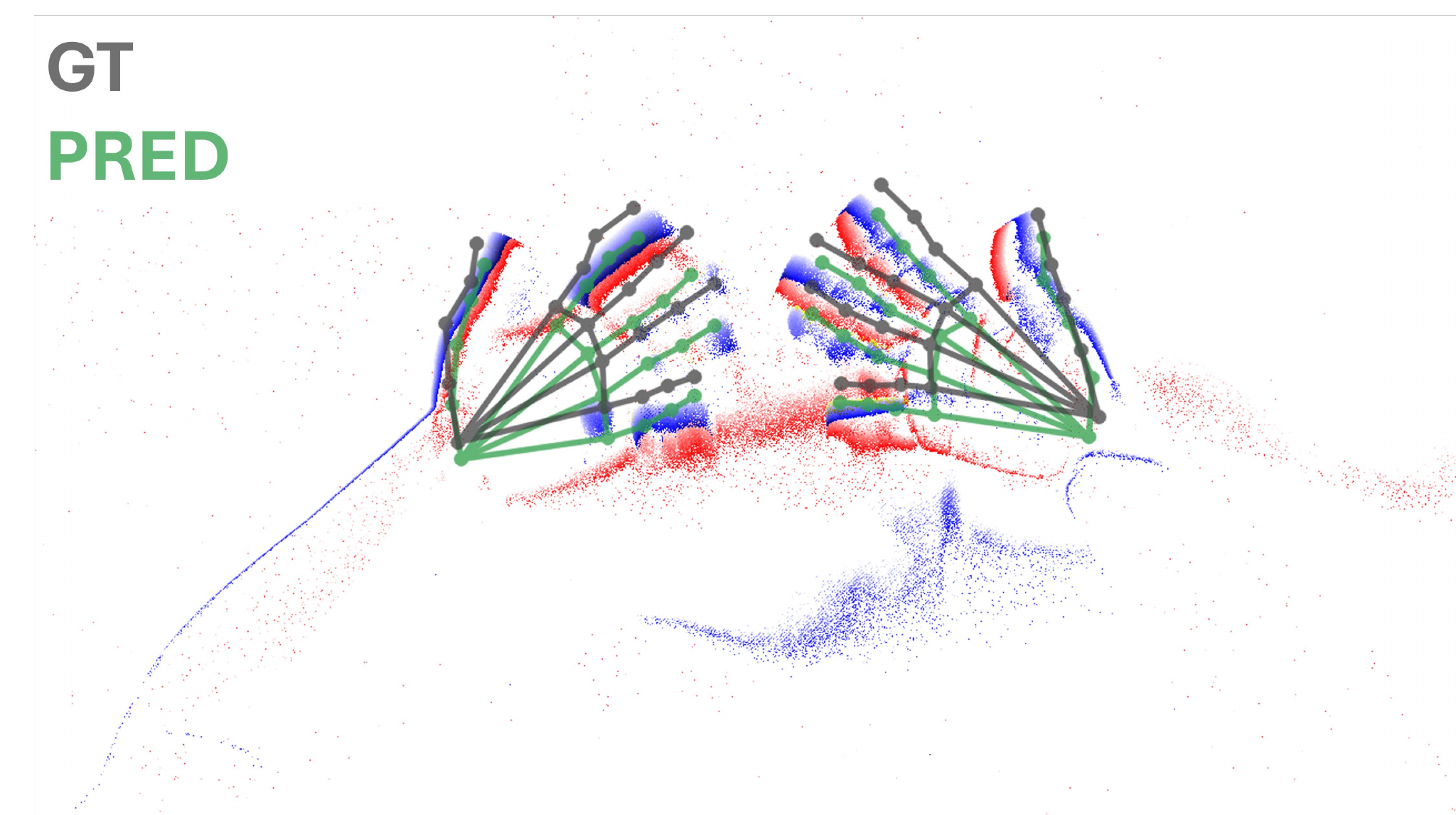}
\caption{Fast-motion scenarios.}
\label{fig:sub_fast}
\end{subfigure}
\hfill
\begin{subfigure}[t]{.24\linewidth}
\centering
\includegraphics[width=\linewidth]{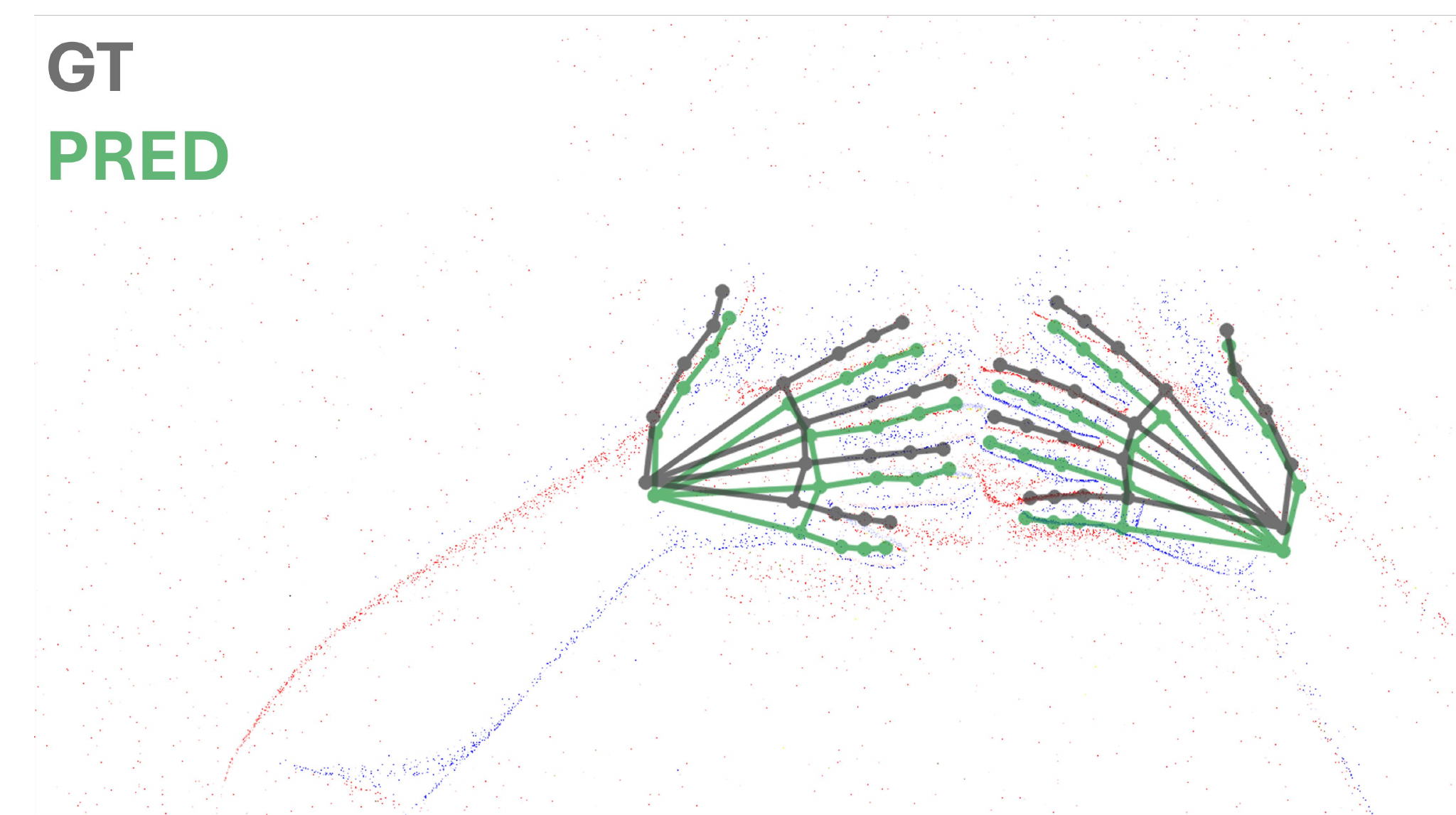}
\caption{Static hand-pose scenarios.}
\label{fig:sub_static}
\end{subfigure}
  \caption{Visualization of EgoEV-HandPose results in four representative scenarios. The examples demonstrate accurate pose reconstruction under normal light, low light, fast motion, and static hand configurations.}
  \label{fig:qualitative_results}
\end{figure}

\textbf{Qualitative Visualization Analysis}: Figure~\ref{fig:qualitative_results} shows accurate reconstructions under normal light, low light, fast motion, and static hand configurations. These examples complement the quantitative results by illustrating the framework's robustness across diverse illumination and motion conditions.

Together, the results in Table~\ref{tab:main_results} establish EgoEV-HandPose as an accurate and computationally efficient framework for egocentric 3D hand pose estimation and gesture recognition. The full framework contains $8.44$\,M parameters and requires $19.86$\,GFLOPs, with a reported throughput of $13.6$\,FPS. This favorable accuracy--complexity trade-off combines robust stereo reconstruction with a compact architecture and provides a strong foundation for future low-latency deployment.

\subsection{Comparison with State-of-the-Art Methods}
\label{sec:comparison}

To evaluate the effectiveness of our proposed framework, we conduct a comprehensive comparison against several representative baselines. These include \textbf{HandMvNet}~\citep{Ali2025HandMvNet}, a state-of-the-art stereo RGB-based method; \textbf{Ev2Hands}~\citep{Millerdurai2024Ev2Hands}, an event-based framework supporting bimanual estimation; \textbf{EvHandPose}~\citep{Jiang2024EvHandPose}, a representative approach for single-hand event-based estimation; \textbf{EventEgo3D}~\citep{millerdurai2024eventego3d}, a monocular event-based approach originally designed for egocentric body pose estimation which we adapted for the hand pose task and \textbf{BEVFormer}~\citep{Li2022BEVFormer}, a heavy-duty multi-view lifting baseline utilizing dense grid construction. 
For a fair comparison, all baselines were re-implemented and fine-tuned on our EgoEVHands dataset. 
HandMvNet uses a grayscale-image input in the adapted evaluation. Ev2Hands downsamples the event streams through convolution to the target resolution, and EvHandPose is trained and evaluated using only right-hand labels because it is designed for single-hand estimation. All adapted baselines use 2D and 3D keypoint supervision rather than MANO-related priors or constraints, providing a unified supervision protocol for comparison on EgoEVHands.

\textbf{Quantitative Analysis on State-of-the-Art Baselines}:
Table~\ref{tab:main_results} demonstrates that EgoEV-HandPose substantially outperforms the RGB-based stereo and event-based baselines in 3D pose estimation. In the HETER low-light subset, our method reduces MPJPE from $82.43$\,mm for HandMvNet to $29.50$\,mm. Overall HETER MPJPE is $30.54$\,mm, compared with $115.13$\,mm for Ev2Hands, and both MPJPE and PA-MPJPE are lower than those of EvHandPose and EventEgo3D. The fixed-backbone comparison and fusion ablations further establish explicit stereo geometry as a principal driver of these gains, beyond the choice of 2D backbone. By anchoring feature fusion to calibrated cross-view correspondences, KeypointBEV directly addresses depth ambiguity and occlusion-induced errors in 3D reconstruction.

\textbf{Model Efficiency}:
The pose-only configuration in Table~\ref{tab:main_results} contains $5.88$\,M parameters and requires $19.86$\,GFLOPs. The full system, including the gesture classifier, contains $8.44$\,M parameters, as detailed in Table~\ref{tab:complexity}. The compact pose network delivers state-of-the-art reconstruction accuracy, while the additional temporal head extends the same geometric representation to gesture recognition.

\textbf{Analysis of 2D-to-3D Lifting Modules}:
To further isolate the contribution of our KeypointBEV module, we conduct an analysis using a fixed 2D backbone across different lifting strategies, as shown in the fixed-backbone rows of Table~\ref{tab:main_results}. The lifting module of EventEgo3D, being monocular, suffers from inherent depth ambiguity, leading to an MPJPE of $51.27$\,mm in the HETER distribution.
Using the same 2D backbone, BEVFormer~\citep{Li2022BEVFormer} obtains $32.03$\,mm MPJPE with $374.14$\,GFLOPs, whereas KeypointBEV obtains $30.54$\,mm with $19.86$\,GFLOPs. Under the HETER cross-subject protocol, KeypointBEV therefore achieves stronger generalization to unseen participants than dense BEV lifting while requiring approximately one nineteenth of the computation. Its sparse, joint-centered representation concentrates modeling capacity on articulated hand geometry, and reprojection-guided sampling supplies explicit cross-view evidence for refinement. These results demonstrate that the geometry-aware design improves both cross-subject pose accuracy and computational efficiency.

\begin{figure}[pos=!htbp,width=\linewidth,align=\centering,abovecap=4pt,belowcap=2pt]
\centering
\begin{subfigure}[t]{0.19\textwidth}
\captionsetup{font=scriptsize,skip=2pt}
    \centering
    \includegraphics[width=\linewidth]{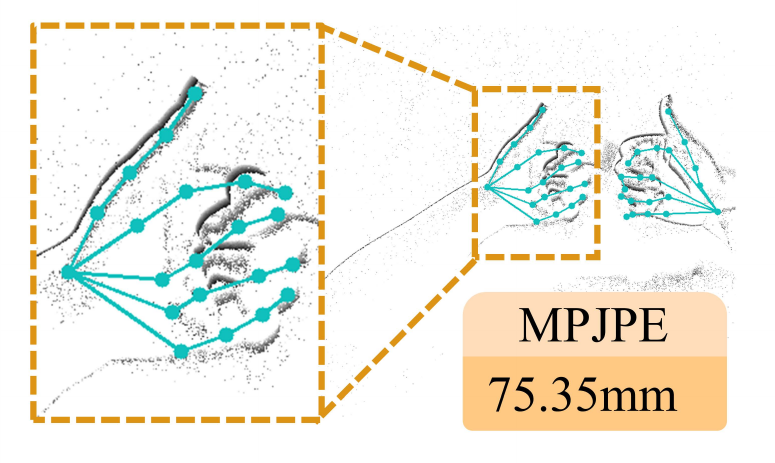}
    \caption{HandMvNet (Case 1)}
\end{subfigure}
\hfill
\begin{subfigure}[t]{0.19\textwidth}
\captionsetup{font=scriptsize,skip=2pt}
    \centering
    \includegraphics[width=\linewidth]{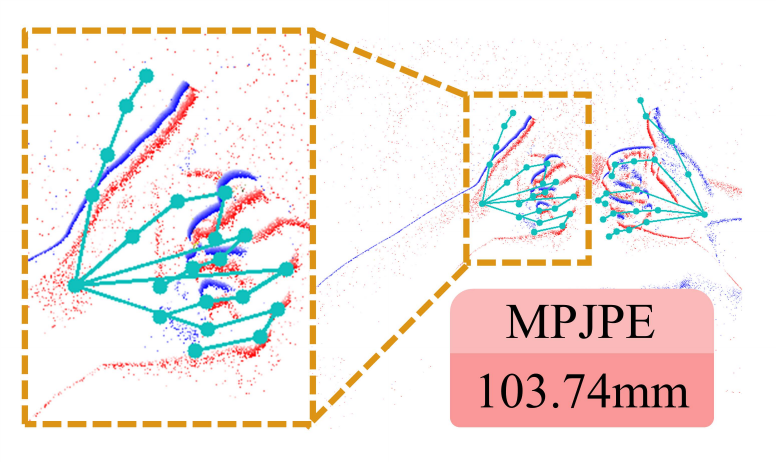}
    \caption{Ev2Hands (Case 1)}
\end{subfigure}
\hfill
\begin{subfigure}[t]{0.19\textwidth}
\captionsetup{font=scriptsize,skip=2pt}
    \centering
    \includegraphics[width=\linewidth]{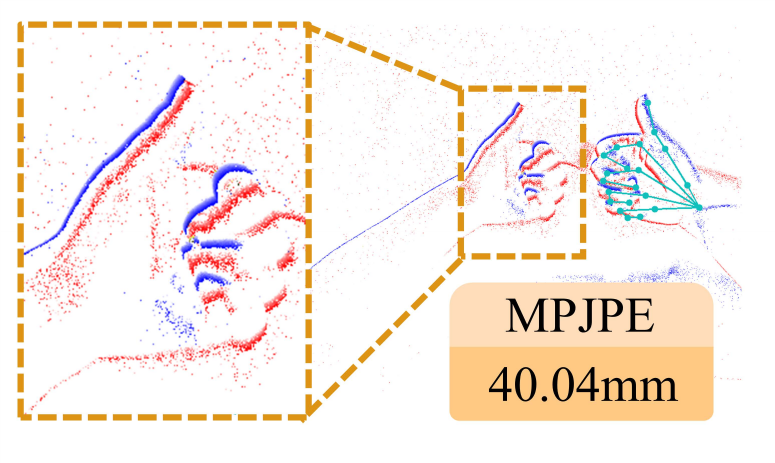}
    \caption{EvHandPose (Case 1)}
\end{subfigure}
\hfill
\begin{subfigure}[t]{0.19\textwidth}
\captionsetup{font=scriptsize,skip=2pt}
    \centering
    \includegraphics[width=\linewidth]{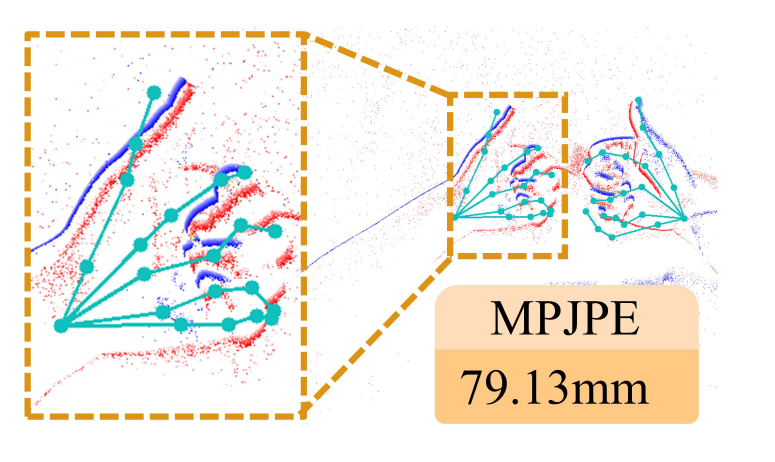}
    \caption{EventEgo3D (Case 1)}
\end{subfigure}
\hfill
\begin{subfigure}[t]{0.19\textwidth}
\captionsetup{font=scriptsize,skip=2pt}
    \centering
    \includegraphics[width=\linewidth]{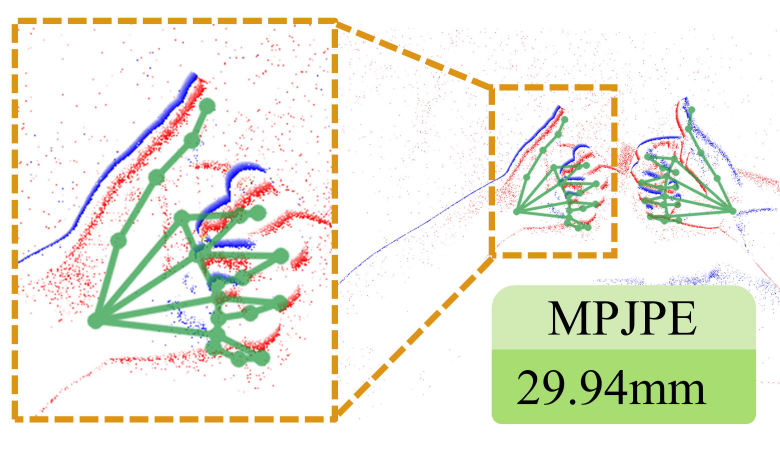}
    \caption{Ours (Case 1)}
\end{subfigure}

\vspace{0.5em}

\begin{subfigure}[t]{0.19\textwidth}
\captionsetup{font=scriptsize,skip=2pt}
    \centering
    \includegraphics[width=\linewidth]{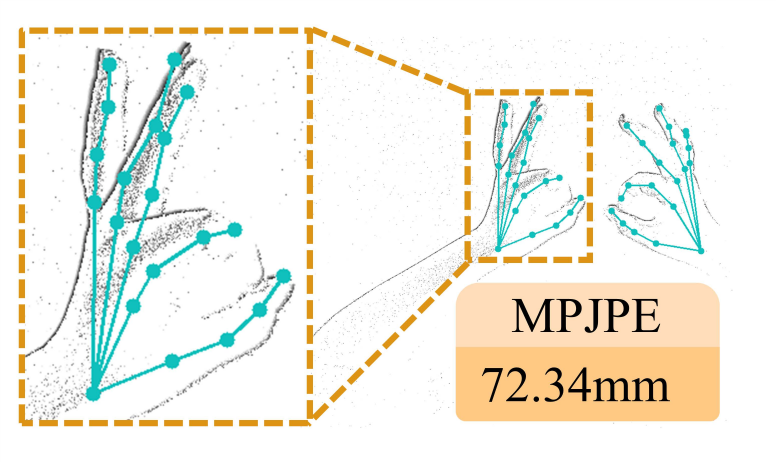}
    \caption{HandMvNet (Case 2)}
\end{subfigure}
\hfill
\begin{subfigure}[t]{0.19\textwidth}
\captionsetup{font=scriptsize,skip=2pt}
    \centering
    \includegraphics[width=\linewidth]{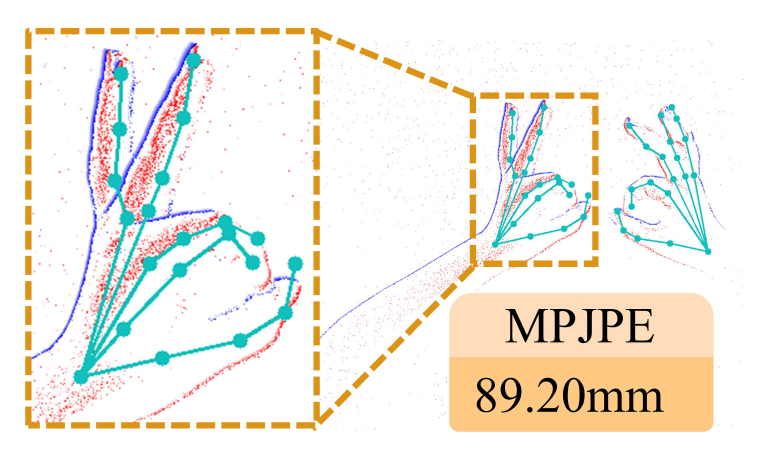}
    \caption{Ev2Hands (Case 2)}
\end{subfigure}
\hfill
\begin{subfigure}[t]{0.19\textwidth}
\captionsetup{font=scriptsize,skip=2pt}
    \centering
    \includegraphics[width=\linewidth]{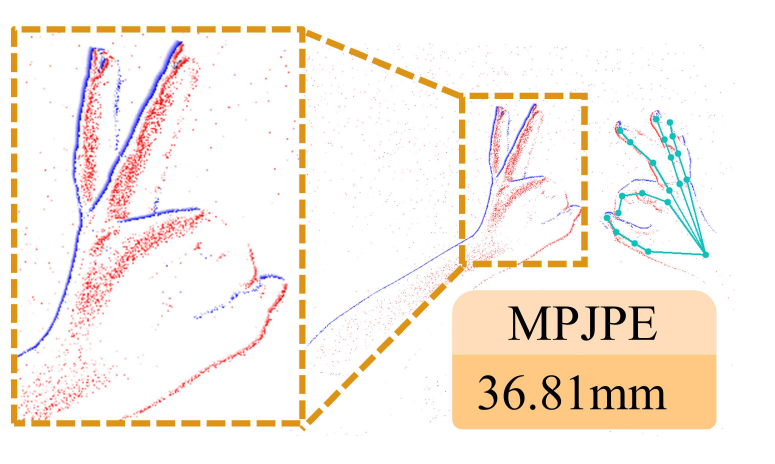}
    \caption{EvHandPose (Case 2)}
\end{subfigure}
\hfill
\begin{subfigure}[t]{0.19\textwidth}
\captionsetup{font=scriptsize,skip=2pt}
    \centering
    \includegraphics[width=\linewidth]{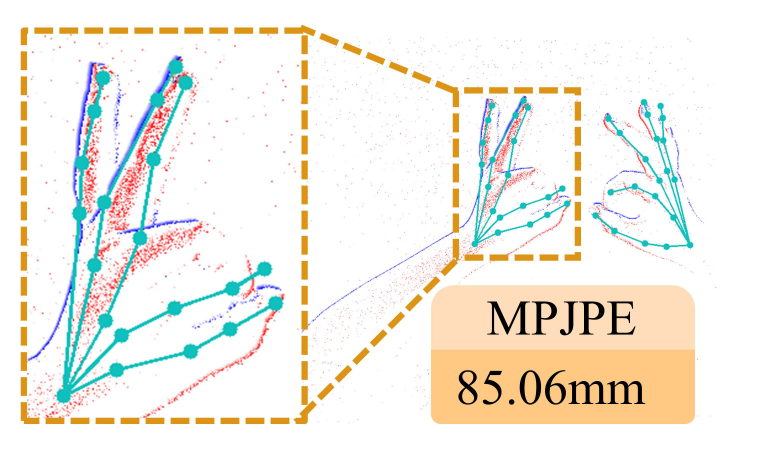}
    \caption{EventEgo3D (Case 2)}
\end{subfigure}
\hfill
\begin{subfigure}[t]{0.19\textwidth}
\captionsetup{font=scriptsize,skip=2pt}
    \centering
    \includegraphics[width=\linewidth]{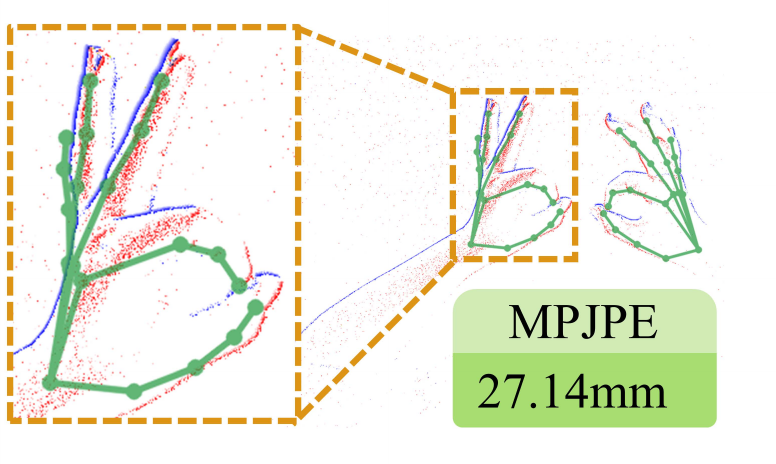}
    \caption{Ours (Case 2)}
\end{subfigure}

\caption{Qualitative comparison between our method and state-of-the-art baselines. The zoomed-in view highlights the superiority of our method in localizing fine-grained finger joints, especially in challenging regions with occlusions and sparse observations.}
\label{fig:sota_comparison}

\end{figure}

\textbf{Qualitative Visualization Analysis}: Fig.~\ref{fig:sota_comparison} provides a visual comparison of our method with the baselines across representative cases. Our stereo configuration and iterative refinement reliably resolve occlusions where monocular baselines fail, especially in challenging bimanual scenarios where EvHandPose is inherently limited by its single-hand design. 

\subsection{Gesture Recognition and Confusion Matrix}
\begin{figure}[pos=!htb,width=\linewidth,align=\centering,abovecap=0pt,belowcap=0pt]
\begin{minipage}[c]{.56\linewidth}
\centering
\includegraphics[width=\linewidth]{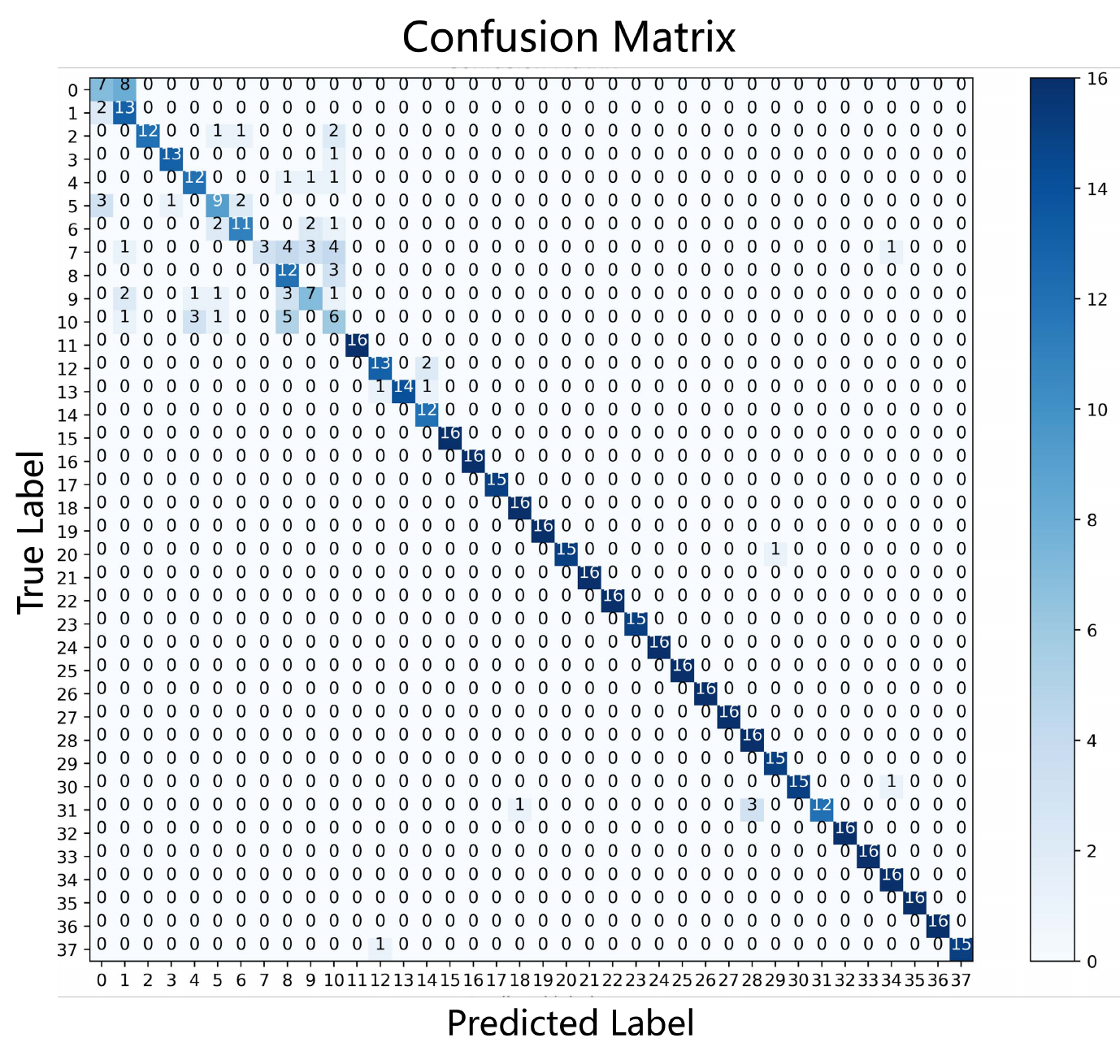}
\end{minipage}\hfill
\begin{minipage}[c]{.40\linewidth}
\LocalFigureWidth
\caption{Confusion matrix of gesture recognition on the EgoEVHands dataset.}
\label{fig:confusion}
\end{minipage}
\end{figure}

Using the refined stereo 2D coordinate sequences described in Sec.~\ref{sec:gesture_recognition}, the jointly optimized framework reaches $86.87\%$ gesture recognition accuracy under HETER. Each frame contributes $168$ coordinates without confidence scores. The confusion matrix in Fig.~\ref{fig:confusion} shows that the pose-based representation effectively distinguishes complex hand interaction categories, including bimanual gestures. These results demonstrate that precise kinematic trajectories provide discriminative spatial and temporal cues that promote high-level gesture understanding. By coupling geometry-aware pose refinement with temporal modeling, EgoEV-HandPose connects accurate 3D hand reconstruction to robust gesture recognition within a unified end-to-end framework.

\subsection{Ablation Study}
\label{sec:ablation}

\begin{figure}[pos=!htb,width=\linewidth,align=\centering,abovecap=4pt,belowcap=2pt]
\begin{minipage}[t]{.48\linewidth}
\vspace{0pt}\centering
\includegraphics[width=\linewidth]{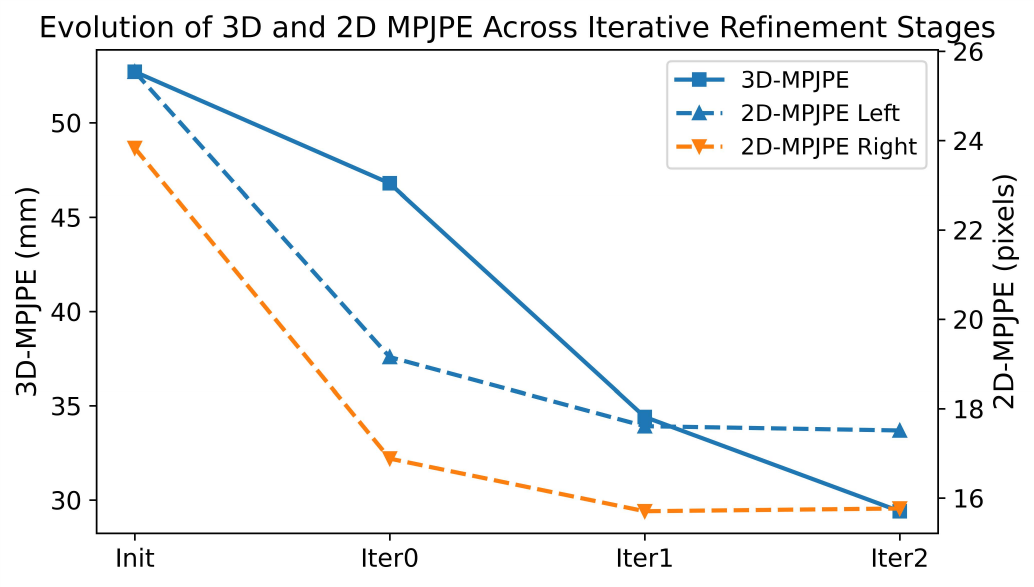}
\LocalFigureWidth
\caption{Evolution of 3D and 2D MPJPE across iterative refinement stages. Left axis: 3D-MPJPE (mm). Right axis: 2D-MPJPE (pixels) for left and right views. The original labels \{Initialization, Iter0, Iter1, Iter2\} correspond to $\{\mathbf{P}^{(0)},\mathbf{P}^{(1)},\mathbf{P}^{(2)},\mathbf{P}^{(3)}\}$: initialization followed by three learned updates.}
\label{fig:mpjpe_iteration}
\end{minipage}\hfill
\begin{minipage}[t]{.48\linewidth}
\vspace{0pt}\centering
\nextfloat
\includegraphics[width=\linewidth]{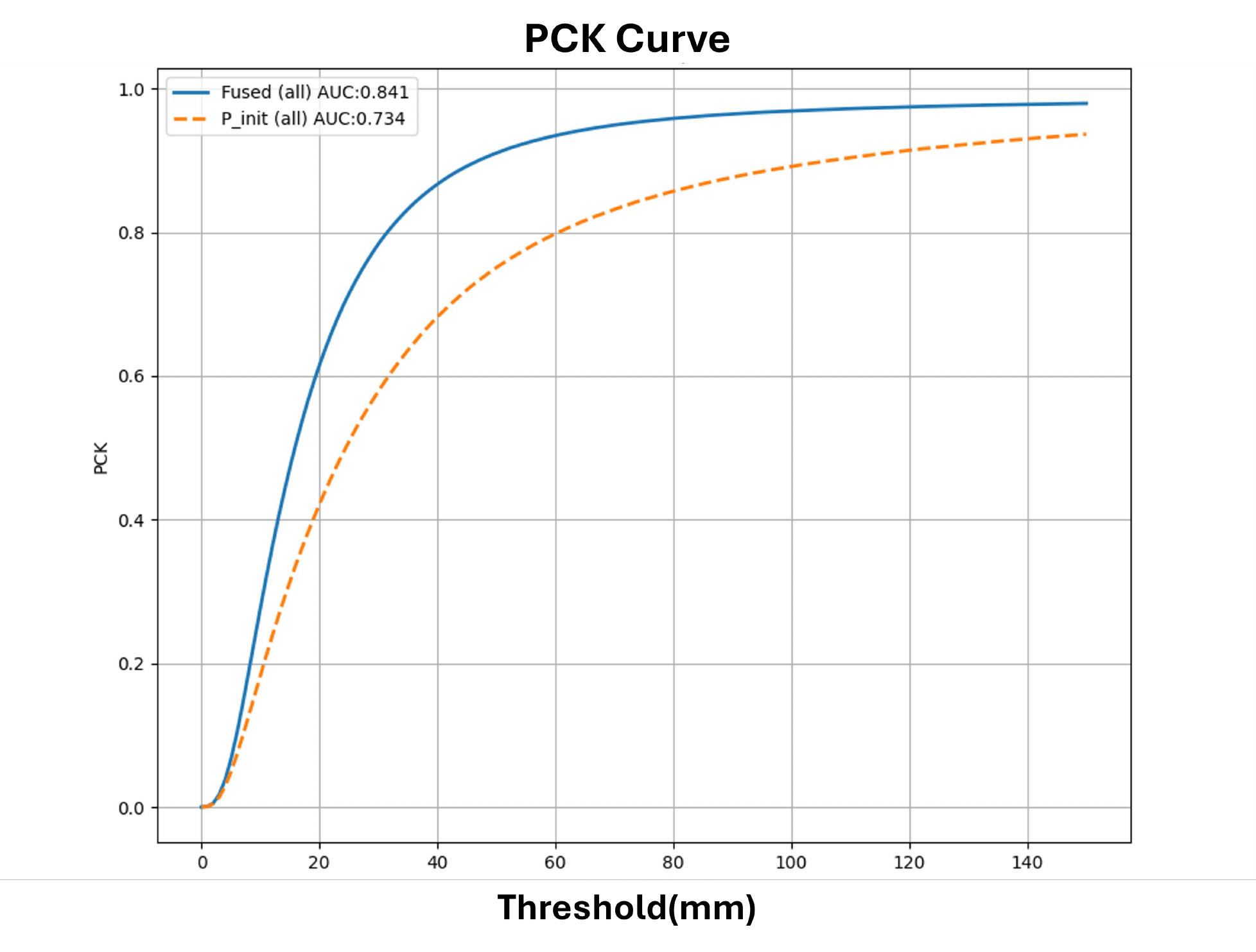}
\LocalFigureWidth
\caption{PCK curve on the full EgoEVHands test set.
The solid blue curve represents the refined prediction and the dashed orange curve represents analytical initialization. The text reports AUC over the $0$--$50$\,mm evaluation interval.}
\label{fig:pck}
\end{minipage}
\end{figure}

To isolate the contributions of the proposed components, specifically the KeypointBEV module, we analyze the performance evolution from the initial coarse stage to the final iteration (see Fig.~\ref{fig:mpjpe_iteration}). 
The reported error decreases from $52.72$\,mm at initialization to $30.54$\,mm after three learned updates. The plot retains its original display labels: Initialization corresponds to $\mathbf{P}^{(0)}$, while Iter0, Iter1, and Iter2 correspond to $\mathbf{P}^{(1)}$, $\mathbf{P}^{(2)}$, and $\mathbf{P}^{(3)}$, respectively.

Fig.~\ref{fig:pck} further illustrates the overall robustness: the fused model achieves a PCK AUC of $0.589$ (thresholds $0 \sim 50$\,mm), substantially outperforming the initial triangulation baseline (AUC $0.436$), indicating that iterative refinement effectively suppresses large localization errors and improves the overall robustness of 3D reconstruction.

We conduct an extensive ablation study to verify the effectiveness of our KeypointBEV module against several alternative fusion strategies, as summarized in Table~\ref{tab:ablation_narrow}. 
Our analysis first investigates the impact of input modalities. 
The comparison between Coarse-3D (Heatmaps) ($112.62\,\text{mm}$) and Coarse-3D (Features) ($87.68\,\text{mm}$) reveals that feature-based representations significantly outperform heatmap-based counterparts. 
This performance gap suggests that feature-level representations preserve richer spatial correspondences than sparse heatmaps, providing a more informative basis for subsequent stereo geometric reasoning.
However, we observe that operating fusion within the 2D domain poses intrinsic limitations for depth estimation in egocentric views. 
As shown in Table~\ref{tab:ablation_narrow}, methods that utilize camera intrinsics and extrinsics to modulate 2D features (PM) still struggle with depth ambiguity. 
Even the Fine-2D variant ($53.49\,\text{mm}$), which learns spatial offsets under parametric guidance, remains constrained by the lack of explicit 3D spatial awareness.
While Cross-attention ($42.98$\,mm) improves feature correspondence across views, it relies primarily on learned statistical matching without explicitly enforcing geometric consistency necessary for precise 3D reconstruction.

KeypointBEV achieves the best performance ($30.54\,\mathrm{mm}$) by explicitly anchoring feature sampling in both views to sparse 3D joint hypotheses and refining these hypotheses with global stereo context. Unlike 2D parameter modulation or unconstrained cross-view matching, this formulation couples visual evidence to calibrated 3D geometry throughout refinement. The ablation results establish the effectiveness of explicit geometric reasoning for resolving depth ambiguity and obtaining accurate 3D hand poses, while sparse joint-centered processing avoids the cost of dense scene representations.

\begin{table}[pos=!htbp,width=\linewidth,abovecap=3pt,belowcap=3pt]
\centering
\caption{Comparison of stereo fusion variants on EgoEVHands.
Mean per-joint position error (MPJPE) is reported in millimeters;
lower is better. The best result is shown in bold.}
\label{tab:ablation_narrow}

\begingroup
\small
\setlength{\tabcolsep}{4pt}
\renewcommand{\arraystretch}{1.08}
\setlength{\extrarowheight}{0pt}
\setlength{\aboverulesep}{2pt}
\setlength{\belowrulesep}{2pt}

\begin{tabular*}{\linewidth}{@{\extracolsep{\fill}}lllr@{}}
\toprule
\textbf{Method} & \textbf{Input} & \textbf{Fusion / prediction} & \textbf{MPJPE (mm)}$\downarrow$ \\
\midrule
Baseline               & 2D joint coordinates      & Triangulation                 & 79.21  \\
Direct-3D              & Heatmaps + camera params. & PM $\rightarrow$ Absolute 3D  & 121.74 \\
Coarse-3D (Heatmaps)    & Heatmaps + camera params. & PM $\rightarrow$ Relative 3D  & 112.62 \\
Coarse-3D (Features)    & Features + camera params. & PM $\rightarrow$ Relative 3D  & 87.68  \\
Fine-2D                & Features + camera params. & PM $\rightarrow$ 2D residuals & 53.49  \\
Fine-2D (Cross-attn.)   & Stereo features           & Cross-view matching           & 42.98  \\
\midrule
KeypointBEV            & 3D joint-query features   & BEV-inspired 3D fusion        & \textbf{30.54} \\
\bottomrule
\end{tabular*}

\par\vspace{2pt}
\begin{minipage}{\linewidth}
\footnotesize\raggedright
PM: parameter modulation; camera params.: camera intrinsics and extrinsics.
The cross-attention variant does not use camera parameters in its matching module.
\end{minipage}

\endgroup
\end{table}

\subsection{Model Complexity Analysis}
\label{sec:complexity}

To assess the practical feasibility of EgoEV-HandPose for egocentric hand perception applications, we provide a detailed breakdown of its computational complexity and parameter distribution in Table~\ref{tab:complexity}.

\begin{table}[pos=!htbp,width=\linewidth]
\centering
\caption{Breakdown of computational complexity and parameter count per module. EgoBlaze parameters are shared between views and counted once, while backbone FLOPs include both stereo forward passes. KeypointBEV uses three refinement updates.}
\label{tab:complexity}
\small
\setlength{\tabcolsep}{10pt}
\renewcommand{\arraystretch}{1.12}
\begin{tabular}{lcc}
\toprule
\textbf{Module} & \textbf{\#Params (M)} & \textbf{FLOPs (G)} \\
\midrule
EgoBlaze & 1.40 & 11.02 ($5.51 \times 2$) \\
KeypointBEV  & 4.48 & 8.84 \\
Action Classifier & 2.56 & 0.003 \\
\midrule
\rowcolor[HTML]{F2F2F2}
\textbf{Total} & \textbf{8.44} & \textbf{19.86} \\
\bottomrule
\end{tabular}

\end{table} 

\textbf{Module-wise Breakdown}: The framework maintains an efficient footprint with a total of $8.44$\,M parameters and $19.86$\,G FLOPs. 
The shared EgoBlaze backbone contains $1.40$\,M parameters, counted once because both views use the same weights. Processing both views requires two forward passes, giving $11.02$\,GFLOPs ($5.51\times2$), without doubling the backbone parameter count. 
The KeypointBEV module, responsible for iterative refinement, constitutes the largest portion of the parameter count ($4.48$\,M). 
Despite accounting for the largest share of the model parameters, the module requires only $8.84$\,GFLOPs, demonstrating that geometry-aware lifting introduces limited computational overhead.
The Action Classifier is designed to be extremely lightweight, contributing negligible computational overhead ($0.003$\,G FLOPs) while processing $T\times168$ refined 2D coordinate sequences for gesture recognition.

\textbf{Accuracy--Complexity Trade-off}:
The fixed-backbone comparison in Table~\ref{tab:main_results} demonstrates a clear accuracy--complexity advantage for sparse, geometry-guided joint refinement. KeypointBEV achieves $30.54$\,mm MPJPE with $19.86$\,GFLOPs, compared with $32.03$\,mm and $374.14$\,GFLOPs for BEVFormer. By focusing computation on articulated joints rather than dense scene grids, the framework delivers stronger cross-subject pose estimation with substantially lower computational cost. This efficient formulation provides a promising basis for low-latency event-based hand perception.

\subsection{Cross-Architecture Generalization on DHP19}
\label{sec:cross_architecture}
To demonstrate the cross-dataset and cross-architecture generalizability and ``plug-and-play'' versatility of KeypointBEV, we extend our evaluation from egocentric hand perception to human pose estimation.
For this purpose, we adopt the DHP19 dataset~\citep{calabrese2019dhp19}, which provides synchronized multi-view event streams for 3D human pose estimation. By capturing data from four Dynamic Vision Sensors (DVS) and providing sub-millimeter 3D ground truth via a Vicon motion capture system, DHP19 serves as a rigorous testbed for evaluating the geometric robustness of event-based reconstruction algorithms in a multi-camera setup.

\begin{figure}[pos=!htbp,width=\linewidth,align=\centering,abovecap=4pt,belowcap=2pt]
\includegraphics[width=.30\linewidth]{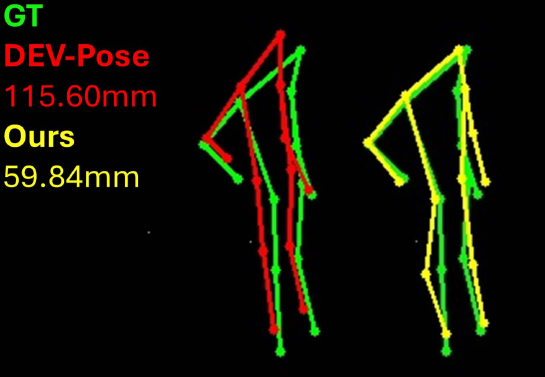}\hspace{.03\linewidth}
\includegraphics[width=.30\linewidth]{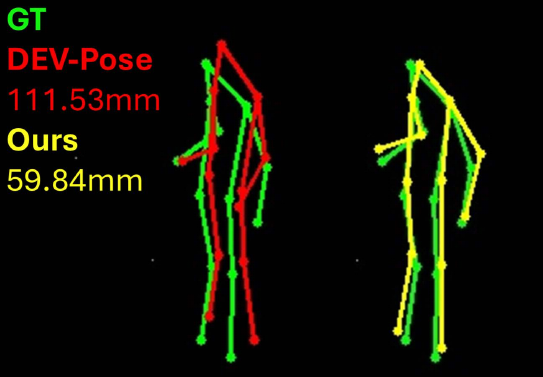}
  \caption{Qualitative comparison on the DHP19 dataset~\citep{calabrese2019dhp19}. Green: GT; Red: DEV-Pose~\citep{yin2023rethinking} baseline; Yellow: Ours (DEV-Pose + KeypointBEV). Our method delivers notably higher pose accuracy and better GT alignment.}
  \label{fig:dhp19_qualitative}
\end{figure}

Specifically, we integrate the KeypointBEV module into the monocular feature backbone of DEV-Pose~\citep{yin2023rethinking}, an event-based human pose estimation framework evaluated on DHP19. 
In this configuration, our module functions as a geometric fusion layer that operates on the learned spatiotemporal features before 3D pose regression. 
As summarized in Table~\ref{tab:dhp19_comparison}, this integration yields a substantial performance improvement: the MPJPE of DEV-Pose (ResNet18) is reduced from $55.53\,mm$ to $47.88\,mm$, a $13.8\%$ reduction. The gain on DHP19, together with the hand-pose results on EgoEVHands, demonstrates that the geometry-aware refinement design generalizes across datasets, backbone architectures, and articulated pose tasks. KeypointBEV thus provides a versatile, plug-and-play geometric fusion component for event-based 3D perception (see Fig.~\ref{fig:dhp19_qualitative}).

\begin{table}[pos=!htbp,width=\linewidth,abovecap=3pt,belowcap=3pt]
\centering
\caption{Comparison of 3D human pose estimation methods on the
DHP19 dataset~\citep{calabrese2019dhp19}.
Baseline results are taken from DEV-Pose~\citep{yin2023rethinking}.
Lower errors indicate better performance.}
\label{tab:dhp19_comparison}

\begingroup
\small
\setlength{\tabcolsep}{10pt}
\renewcommand{\arraystretch}{1.12}
\setlength{\extrarowheight}{0pt}
\setlength{\aboverulesep}{2pt}
\setlength{\belowrulesep}{2pt}

\begin{tabular}{@{}lrr@{}}
\toprule
\textbf{Method}
& \textbf{2D MPJPE (px)} $\downarrow$
& \textbf{3D MPJPE (mm)} $\downarrow$ \\
\midrule
DHP19~\citep{calabrese2019dhp19}
& 7.67 & 87.90 \\

MobileHumanPose-S~\citep{choi2021mobilehumanpose}
& 5.65 & 64.14 \\

Pose-ResNet18~\citep{xiao2018simple}
& 5.37 & 61.03 \\

Pose-ResNet50~\citep{xiao2018simple}
& 5.28 & 59.83 \\

PointNet~\citep{qi2017pointnet}
& 7.29 & 82.46 \\

DGCNN~\citep{wang2019dynamic}
& 6.83 & 77.32 \\

Point Transformer~\citep{zhao2021point}
& 6.46 & 73.37 \\

VMV-PointTrans~\citep{xie2022vmv}
& 9.13 & 103.23 \\

VMST-Net~\citep{liu2023voxel}
& 6.45 & 73.07 \\

DEV-Pose (DHP19 backbone)~\citep{yin2023rethinking}
& 6.27 & 71.01 \\

DEV-Pose (ResNet18)~\citep{yin2023rethinking}
& 4.93 & 55.53 \\

\midrule

Ours (DEV-Pose + KeypointBEV)
& \textbf{4.79} & \textbf{47.88} \\

\bottomrule
\end{tabular}
\endgroup
\end{table}

\section{Conclusion}
\label{sec:conclusion}
We presented EgoEV-HandPose, a robust and computationally efficient end-to-end framework for egocentric 3D bimanual hand pose estimation and gesture recognition from stereo events. Its KeypointBEV module combines sparse 3D joint hypotheses with reprojection-guided stereo fusion and iterative refinement to resolve depth ambiguity and improve geometric consistency under occlusion. We also introduced EgoEVHands, the first large-scale real-world stereo event dataset for egocentric 3D bimanual hand pose estimation and gesture recognition, extending EgoEvGesture recordings with dense pose-level supervision and comprising 2,498 annotated stereo sequence pairs across 38 gesture categories. Extensive experiments demonstrate that EgoEV-HandPose achieves state-of-the-art performance on EgoEVHands, maintaining exceptional robustness under low illumination and severe self- and mutual occlusions. Fixed-backbone comparisons and fusion ablations establish explicit stereo geometry as a principal source of the performance gains, while the refined kinematic trajectories provide discriminative cues for accurate gesture recognition. KeypointBEV further achieves stronger cross-subject generalization than dense BEV lifting at substantially lower computational cost. Its successful integration with DEV-Pose on DHP19 demonstrates the strong cross-dataset and cross-architecture generalizability and plug-and-play versatility of the geometric refinement design. Together, these results establish geometry-aware stereo event fusion as an effective approach to robust and efficient egocentric hand perception.

Future work will expand subject diversity and incorporate self-supervised learning to reduce dependence on manual annotations. We will investigate adaptive structural gating to accommodate single-hand and bimanual interactions, and explore one-shot geometry-aware architectures and tracking-based optimization to reduce redundant frame-by-frame refinement. To enable ultra-low-latency deployment on low-power wearable devices, we will optimize temporal feature reuse and evaluate end-to-end latency and power consumption on target hardware. Deploying this perception pipeline in real-time AR/VR and wearable interaction systems remains an important direction for translating accurate stereo event-based hand perception into practical human--computer interaction.

\FloatBarrier
\printcredits
\IfFileExists{paper.bib}{%
  \bibliographystyle{cas-model2-names}
  \bibliography{paper}
}{%

}
\end{document}